\pdfoutput=1

\documentclass[10pt,twocolumn,letterpaper]{article}

\usepackage[pagenumbers]{cvpr} 

\usepackage{graphicx}
\usepackage{amsmath}
\usepackage{amssymb}
\usepackage{booktabs}
\usepackage{float}
\usepackage{algorithm}
\usepackage{algorithmic}

\usepackage[pagebackref,breaklinks,colorlinks]{hyperref}

\usepackage[capitalize]{cleveref}
\crefname{section}{Sec.}{Secs.}
\Crefname{section}{Section}{Sections}
\Crefname{table}{Table}{Tables}
\crefname{table}{Tab.}{Tabs.}

\def\cvprPaperID{7433} 
\def\confName{CVPR}
\def\confYear{2023}

\begin{document}

\title{ContraNeRF: Generalizable Neural Radiance Fields for Synthetic-to-real \\ Novel View Synthesis via Contrastive Learning}

\author{Hao Yang{$^1$}
~~~~ Lanqing Hong{$^2$}
~~~~ Aoxue Li{$^2$}
~~~~ Tianyang Hu{$^2$}\\
~~~~ Zhenguo Li{$^2$}
~~~~ Gim Hee Lee{$^3$}\footnotemark[1] 
~~~~ Liwei Wang{$^{1,4}$}\footnotemark[1] \\
{\normalsize
{$^1$}Center for Data Science, Peking University
~~ {$^2$}Huawei, China
~~ {$^3$}School of Computing, National University of Singapore
}\\
{\normalsize
{\hspace*{-12pt}}
}
{\normalsize
{$^4$}National Key Laboratory of General Artificial Intelligence, School of Intelligence Science and Technology, Peking University
}\\
{\tt\small \{haoy@stu, wanglw@cis\}.pku.edu.cn
~~ gimhee.lee@comp.nus.edu.sg}\\
{\tt\small \{honglanqing, liaoxue2, hutianyang1, li.zhenguo\}@huawei.com  }
}
\maketitle
\renewcommand{\thefootnote}{\fnsymbol{footnote}}
\footnotetext[1]{Joint last authorship.}

\begin{abstract}

Although many recent works have investigated generalizable NeRF-based novel view synthesis for unseen scenes, they seldom consider the synthetic-to-real generalization, which is desired in many practical applications.
In this work, we first investigate the effects of synthetic data in synthetic-to-real novel view synthesis and surprisingly observe that models trained with synthetic data tend to produce sharper but less accurate volume densities.
For pixels where the volume densities are correct, fine-grained details will be obtained.
Otherwise, severe artifacts will be produced.
To maintain the advantages of using synthetic data while avoiding its negative effects, we propose to introduce geometry-aware contrastive learning to learn multi-view consistent features with geometric constraints. 
Meanwhile, we adopt cross-view attention to further enhance the geometry perception of features by querying features across input views.
Experiments demonstrate that under the synthetic-to-real setting, our method can render images with higher quality and better fine-grained details, outperforming existing generalizable novel view synthesis methods in terms of PSNR, SSIM, and LPIPS.
When trained on real data, our method also achieves state-of-the-art results.
\href{https://contranerf.github.io/}{https://contranerf.github.io/}
\end{abstract}

\section{Introduction}
\label{sec:intro}

Novel view synthesis is a classical problem in computer vision, which aims to produce photo-realistic images for unseen viewpoints\cite{chen1993view, buehler2001unstructured, waechter2014let, debevec1996modeling, wood2000surface}. 
Recently, Neural Radiance Fields (NeRF)\cite{martin2021nerf} proposes to achieve novel view synthesis through continuous scene modeling through a neural network, which quickly attracts widespread attention due to its surprising results.
However, the vanilla NeRF is actually designed to fit the continuous 5D radiance field of a given scene, which often fails to generalize to new scenes and datasets.
How to improve the generalization ability of neural scene representation is a challenging problem.

\begin{figure*}
  \centering
  \begin{subfigure}{0.19\linewidth}
    \includegraphics[width=\textwidth]{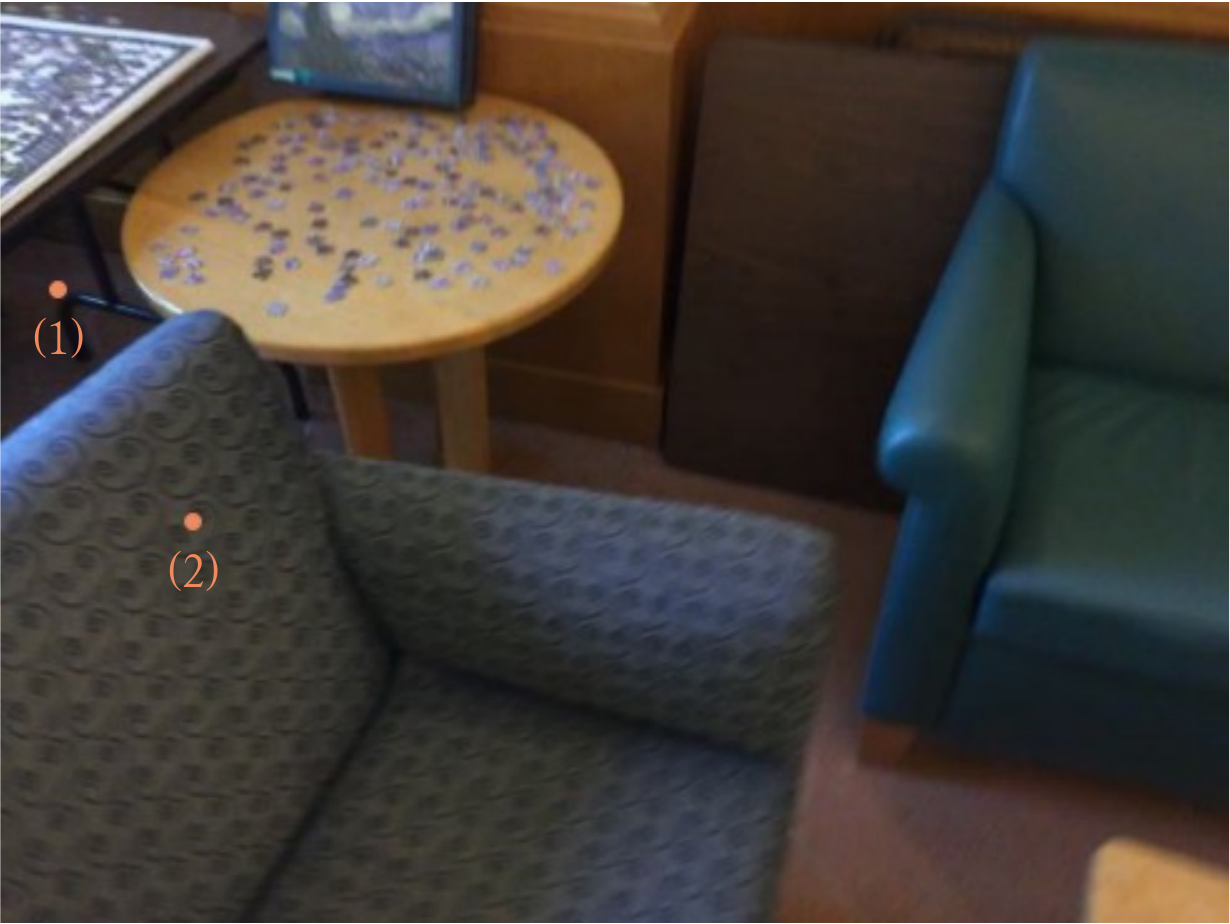}
    \caption{Ground truth}
    \label{fig:gt1}
  \end{subfigure}
  \begin{subfigure}{0.19\linewidth}
    \includegraphics[width=\textwidth]{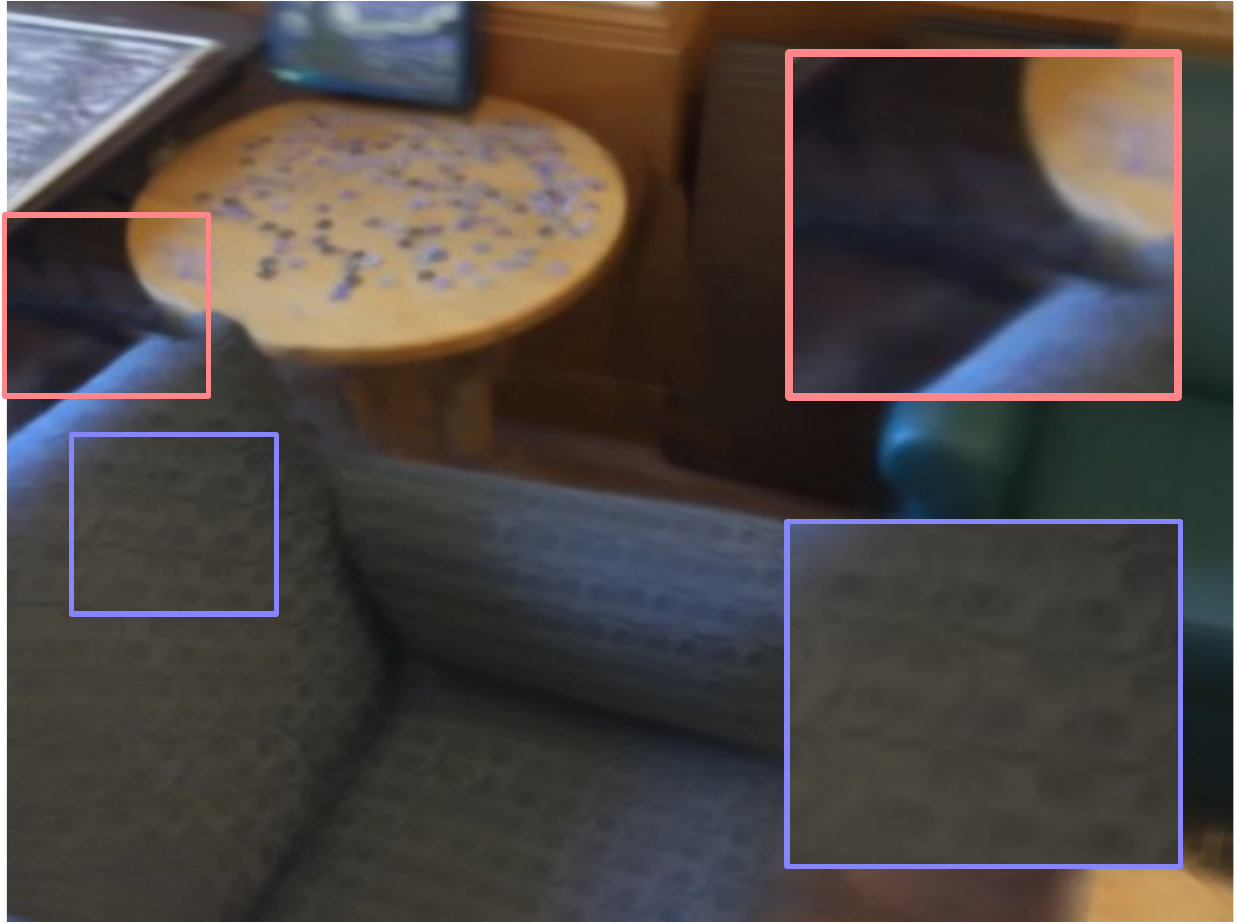}
    \caption{Pred (ScanNet)}
    \label{fig:pred-sacnnet}
  \end{subfigure}
  \begin{subfigure}{0.19\linewidth}
    \includegraphics[width=\textwidth]{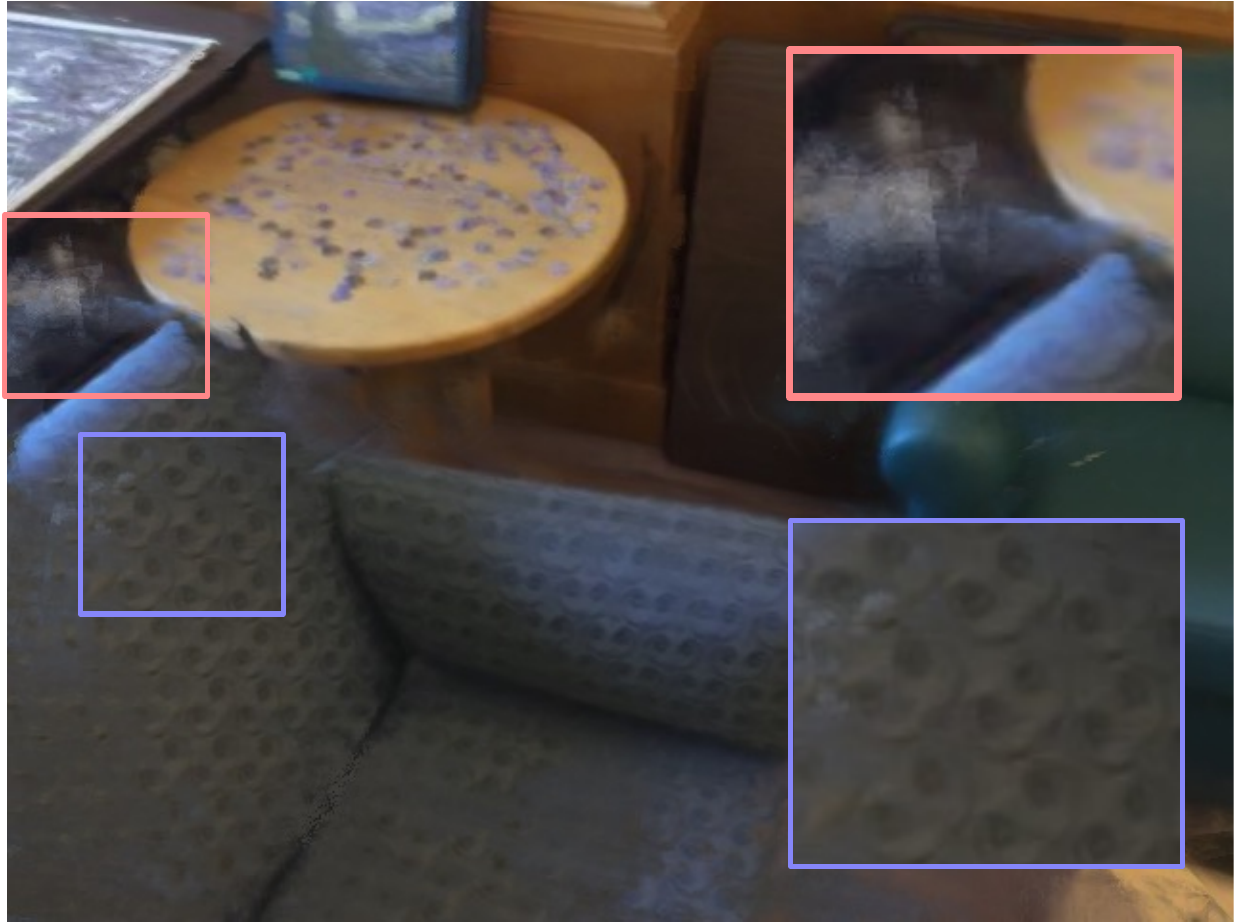}
    \caption{Pred (3D-FRONT)}
    \label{fig:pred-3dfront}
  \end{subfigure}
  \begin{subfigure}{0.19\linewidth}
    \includegraphics[width=\textwidth]{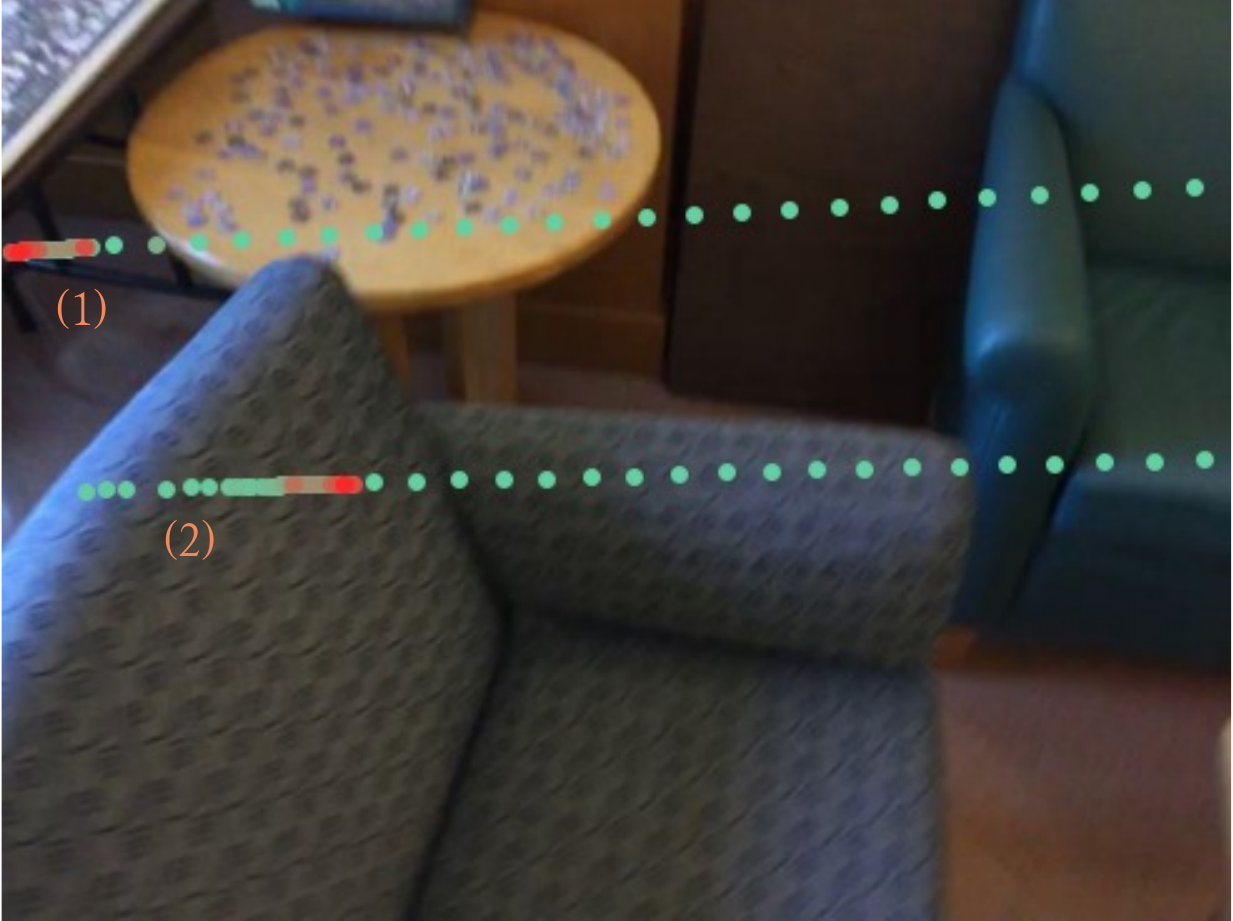}
    \caption{Density (ScanNet)}
    \label{fig:den-scannet}
  \end{subfigure}
  \begin{subfigure}{0.19\linewidth}
    \includegraphics[width=\textwidth]{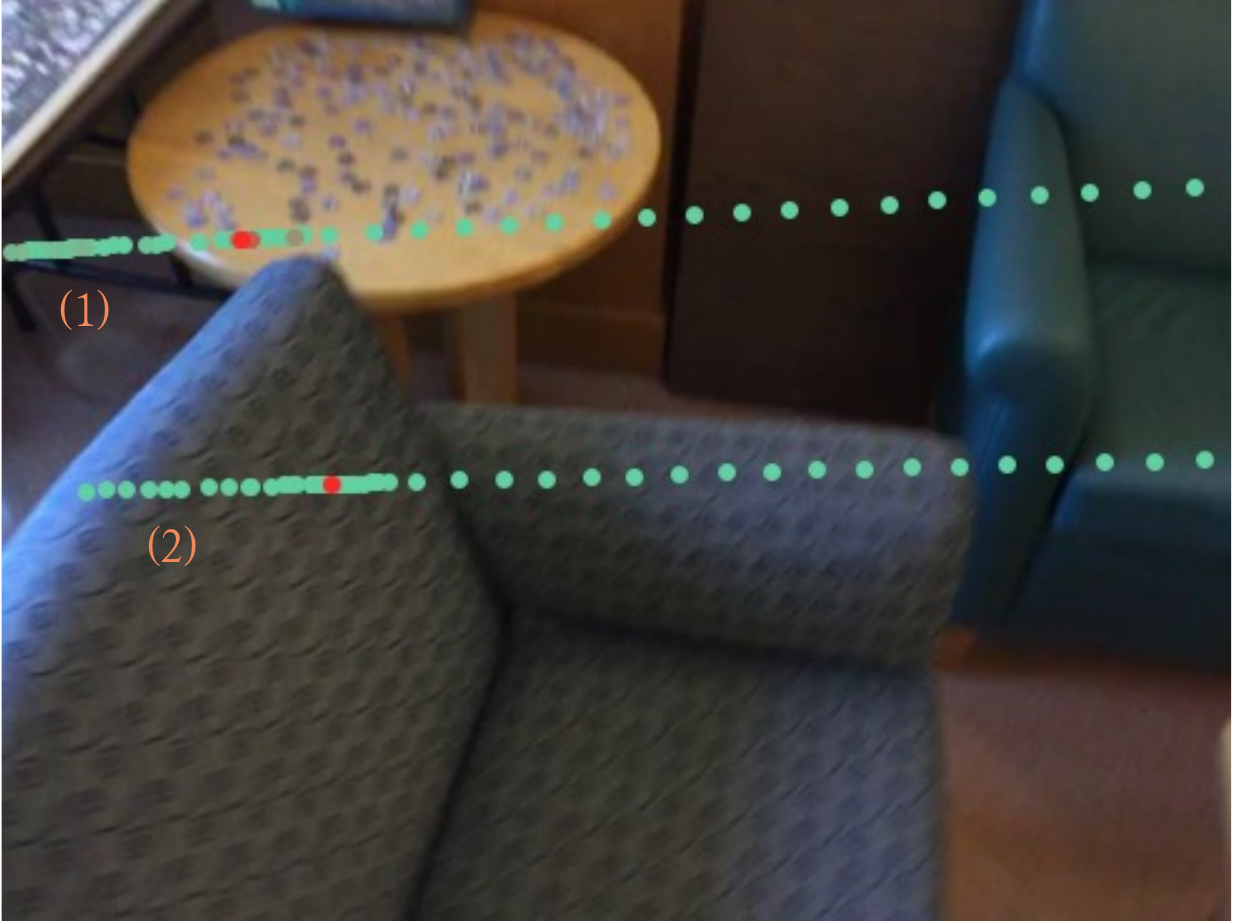}
    \caption{Density (3D-FRONT)}
    \label{fig:den-3dfront}
  \end{subfigure}
  \caption{Fig.\ref{fig:gt1} is the ground truth of the target image to be rendered. Fig.\ref{fig:pred-sacnnet} and Fig.\ref{fig:pred-3dfront} are the rendered images when models are trained on ScanNet\cite{dai2017scannet} and 3D-FRONT\cite{fu20213d}, respectively. Compare to Fig.\ref{fig:pred-sacnnet}, Fig.\ref{fig:pred-3dfront} is more detailed (see purple box) but has more artifacts (see pink box). Fig.\ref{fig:den-scannet} and Fig.\ref{fig:den-3dfront} further show the volume density (the redder the color, the higher the density) along the epipolar line projected from the orange points in Fig.\ref{fig:gt} to the source view. The model trained on 3D-FRONT prefers to predict the volume density with a sharper distribution, but sometimes the predicted volume density is not accurate (see line 1 in Fig.\ref{fig:den-3dfront}), resulting in severe artifacts.}
  \label{fig:preliminary}
\end{figure*}

\begin{figure}[!tb]
    \centering
    \begin{subfigure}{0.49\linewidth}
        \includegraphics[width=\textwidth]{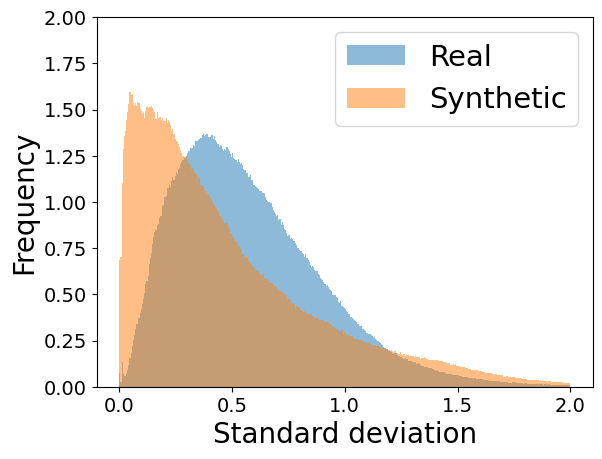}
        \caption{Deviation.}
        \label{fig:var}
    \end{subfigure}
    \begin{subfigure}{0.49\linewidth}
        \includegraphics[width=\textwidth]{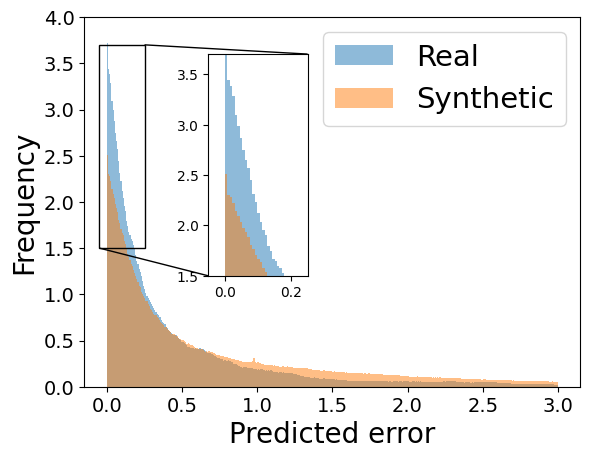}
        \caption{Error.}
        \label{fig:err}
    \end{subfigure}
    \caption{\textbf{Deviation and error of predicted depth when trained with synthetic and real data, respectively.} 
    We count the deviation and error of the predicted depth for each pixel in the test dataset and plot them as the histogram.
    The depth is calculated by aggregating depth of the sampled points along the rendering ray, similar to the process of color rendering.
    Compared to the model trained with the real data, the model trained with the synthetic data tends to predict depths with small deviations but large errors, i.e., density distributions that are sharper but less geometrically accurate.}
    \label{fig:depth}
\end{figure}

Recent works, such as pixelNeRF\cite{yu2021pixelnerf}, IBRNet\cite{wang2021ibrnet}, MVSNeRF\cite{chen2021mvsnerf} and GeoNeRF\cite{johari2022geonerf}, investigate how to achieve generalizable novel view synthesis based on neural radiance fields.
However, these works mainly focus on the generalization of NeRF to unseen scenes and seldom consider the synthetic-to-real generalization, i.e., training NeRF with synthetic data while testing it on real data.
On the other hand, synthetic-to-real novel view synthesis is desired in many practical applications where the collection of dense view 3D data is expensive (e.g., autonomous driving, robotics, and unmanned aerial vehicle\cite{truong2021bi}).
Although some works directly use synthetic data such as Google Scanned Objects~\cite{downs2022google} in model training, they usually overlook the domain gaps between the synthetic and real data as well as possible negative effects of using synthetic data.
In 2D computer vision, it is common sense that synthetic training data usually hurts the model's generalization ability to real-world applications~\cite{yasarla2020syn2real,chen2021psd,bonatti2020learning}.
\emph{Will synthetic data be effective in novel view synthesis?}

In this work, we first investigate the effectiveness of synthetic data in NeRF's training via extensive experiments. Specifically, we train generalizable NeRF models using a synthetic dataset of indoor scenes called 3D-FRONT\cite{fu20213d}, and test the models on a real indoor dataset called ScanNet\cite{dai2017scannet}. 
Surprisingly, we observe that the use of synthetic data tends to result in more artifacts on one hand but better fine-grained details on the other hand (see Fig.\ref{fig:preliminary} and Sec.\ref{sec:Syn2Real} for more details).
Moreover, we observe that models trained on synthetic data tend to predict sharper but less accurate volume densities (see Fig.\ref{fig:depth}).
In this case, better fine-grained details can be obtained once the prediction of geometry (i.e., volume density) is correct, while severe artifacts will be produced otherwise.
This motivates us to consider one effective way to generalize from synthetic data to real data in a geometry-aware manner.

\begin{figure*} 
    \centering
    \includegraphics[width=0.99\textwidth]{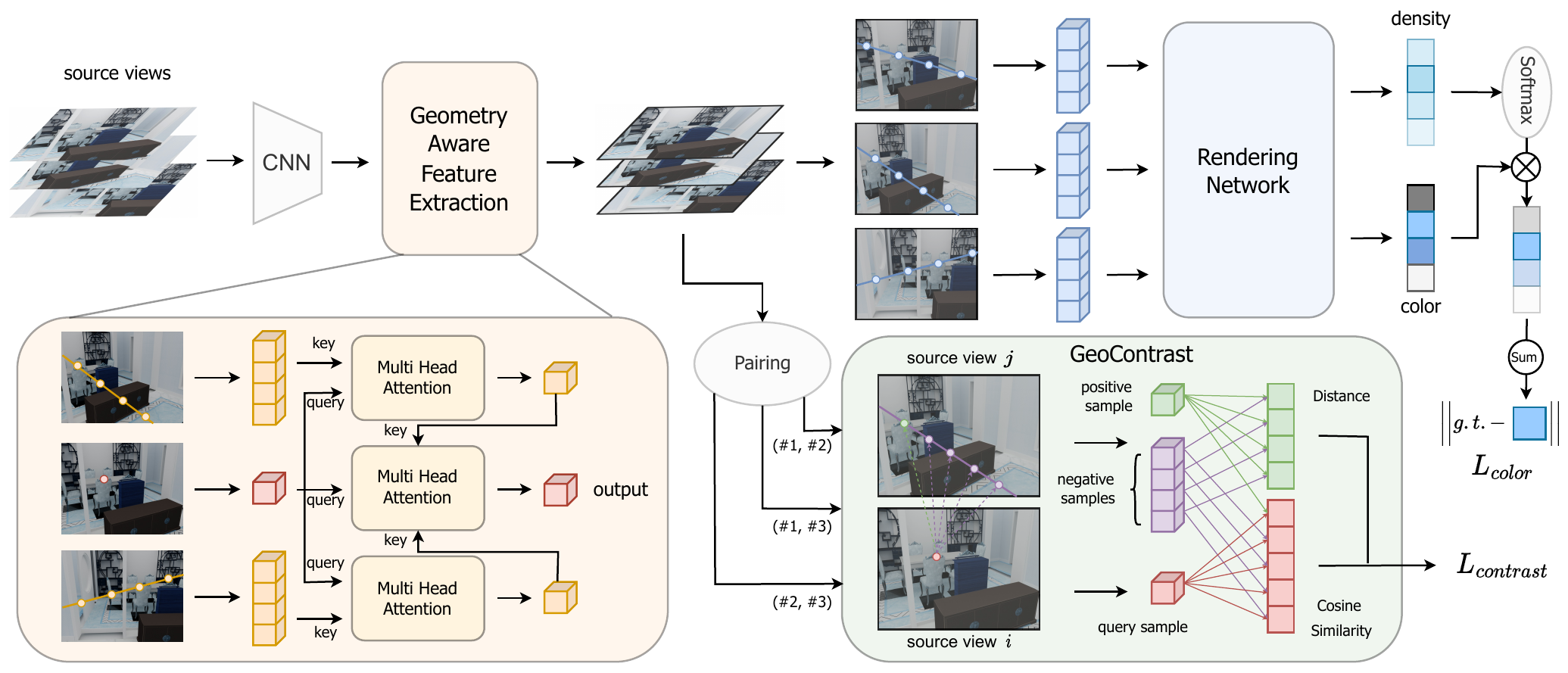}
    \caption{\textbf{Pipeline of our ContraNeRF}. 1) We first use a shared CNN to extract features for input source views. Then for each source view, we query features from other source views and aggregate them to get the geometrically enhanced feature maps (Sec.\ref{sec:crossgam}). 2) For each pair of source views, we compute the contrastive loss using our GeoContrast (Sec.\ref{sec:geocontrast}). Specifically, for the pixel in the $i$-th source view, we project it to the $j$-th source view and sample a collection of projections to get positive and negative samples. Then the weighted contrastive loss is calculated by considering the distance between the positive sample and the negative samples. 3) Finally, for each ray in the target view, we compute colors and densities for a set of samples along the ray by aggregating local features from source views, and accumulate colors and densities to render images (Sec.\ref{sec:render}).}  
    \label{fig:pipeline}
\end{figure*}

To improve the synthetic-to-real generalization ability of NeRF, we propose ContraNeRF, a novel approach that generalizes well from synthetic data to real data via contrastive learning with geometry consistency.
In many 2D vision tasks, contrastive learning has been shown to improve the generalization ability of models~\cite{yang2022towards,yao2022pcl,kim2021selfreg} by enhancing the consistency of positive pairs.
In 3D scenes, geometry is related to multi-view appearance consistency\cite{wang2021ibrnet, liu2022neural}, and contrastive learning may help models predict accurate geometry by enhancing multi-view consistency.
In this paper, we propose geometry-aware contrastive learning to learn a multi-view consistent features representation by comparing the similarities of local features for each pair of source views (see Fig.\ref{fig:pipeline}).
Specifically, for pixels of each source view, we first aggregate information along the ray projected to other source views to get the geometry-enhanced features.
Then, we sample a batch of target pixels from each source view as the training batch for contrastive learning and project them to other views to get positive and negative samples.
The InfoNCE loss\cite{oord2018representation} is calculated in a weighted manner.
Finally, we render the ray by learning a general view interpolation function following \cite{wang2021ibrnet}.
Experiments show that when trained on the synthetic data, our method outperforms the recent concurrent generalizable NeRF works~\cite{yu2021pixelnerf, wang2021ibrnet, chen2021mvsnerf, johari2022geonerf, liu2022neural} and can render high-quality novel view while preserving fine-grained details for unseen scenes.
Moreover, under the real-to-real setting, our method also performs better than existing neural radiance field generalization methods.
In summary, our contributions are:
\begin{enumerate}
    \item Investigate the effects of synthetic data in NeRF-based novel view synthesis and observe that models trained on synthetic data tend to predict sharper but less accurate volume densities when tested on real data;
    \item Propose geometry-aware contrastive learning to learn  multi-view consistent features with geometric constraints, which significantly improves the model's synthetic-to-real generalization ability;
    \item Our method achieves state-of-the-art results for generalizable novel view synthesis under both synthetic-to-real and real-to-real settings.
\end{enumerate}

\section{Related work}
\label{sec:related-work}

\noindent \textbf{NeRF generalization.} Recently, we have witnessed a major breakthrough in novel view synthesis by NeRF\cite{mildenhall2021nerf} and the following works\cite{yu2021plenoctrees, pumarola2021d, li2021neural, martin2021nerf}.
However, these methods can only be applied to a single scene and cannot be generalized to unseen scenes.
Therefore, generalization NeRF\cite{yu2021pixelnerf, wang2021ibrnet, reizenstein2021common, chen2021mvsnerf, johari2022geonerf, liu2022neural, zhang2022nerfusion} has subsequently become a hot research direction which aims to construct a neural radiance field on-the-fly using only a few images as input.
IBRNet\cite{wang2021ibrnet} uses a similar network but it synthesizes novel views by blending pixels from nearby views with weights and volume densities inferred by a network comprising an MLP and ray transformer.
Neuray\cite{liu2022neural} further considers the visibility of each nearby view when constructing radiance fields and achieves good performance.
MVSNeRF\cite{chen2021mvsnerf} and GeoNeRF\cite{johari2022geonerf} leverage deep MVS techniques to achieve across-scene neural radiance field estimation for high-quality view synthesis.
However, these works overlook the possible negative effects of using synthetic data and have difficulty in generalizing from synthetic data to real data. 

\noindent \textbf{NeRF with Geometry.}
Some recent work attempts to introduce geometry information into NeRF's training.
NerfingMVS\cite{wei2021nerfingmvs} uses the depth priors to guide the optimization process of NeRF\cite{mildenhall2021nerf}.
DS-NeRF\cite{deng2022depth} explores depth as additional supervision to guide the geometry learned by NeRF.
RegNeRF\cite{niemeyer2022regnerf} regularizes the geometry and appearance of patches rendered from unobserved viewpoints in sparse input scenarios.
However, all these methods are designed for single-scene reconstruction without generalization ability.

\noindent \textbf{Contrastive learning.} Contrastive learning is a prevailing self-supervised learning technique~\cite{he2020momentum,chen2020improved,misra2020self,tian2020makes}, which proposes to construct supervision information by treating each image as a class, training a model by pulling positive sample pairs closer while pushing negative sample pairs away with InfoNCE loss~\cite{oord2018representation}.
Compared with traditional supervised learning, contrastive learning has been shown to have better generalization ability for various 2D vision tasks~\cite{yang2022towards,yao2022pcl,kim2021selfreg}. 
Previous works \cite{chen2020simple, he2020momentum, chen2020improved} usually take contrastive learning in an instance-level manner. Some recent works \cite{xie2021propagate, wang2021dense, hu2021region} try to apply contrastive learning at the pixel level for learning dense feature representations.

\noindent \textbf{Synthetic-to-real generalization} is a long-standing task that is desired in many applications, including autonomous driving, robotics, and unmanned aerial vehicle~\cite{truong2021bi}.
Although many works have considered Synthetic-to-real transfer for many tasks, such as classification~\cite{kumar2020syn2real}, object detection~\cite{cai2019exploring}, image deraining~\cite{yasarla2020syn2real}, and pose estimation~\cite{doersch2019sim2real},
these methods cannot be directly applied to novel view synthesis.
The synthetic-to-real generalization of neural radiance based novel view synthesis is seldom explored.


\section{Problem Formulation}
\label{sec:problem-formulation}

\subsection{Generalizable Neural Radiance Fields}

In this section, we first introduce the setting of Generalizable Neural Radiance Fields~\cite{yu2021pixelnerf, wang2021ibrnet, reizenstein2021common, chen2021mvsnerf, johari2022geonerf, liu2022neural, zhang2022nerfusion}.
Let $\mathcal{D}_{train} = \{ \mathcal{I}_i, \mathcal{K}_i, \mathcal{E}_i \}_{i=1}^{N_{train}}$ denote the training set, where $\mathcal{I}_i = \{ \mathbf{I}_j \}_{j=1}^{N_{view}^i}$, $\mathcal{K}_i = \{ \mathbf{K}_j \}_{j=1}^{N_{view}^i}$, $\mathcal{E}_i = \{ \mathbf{E}_j=[\mathbf{R}_j, \mathbf{t}_j] \}_{j=1}^{N_{view}^i}$ are the images, camera intrinsic and extrinsic parameters of the $i$-th scenes respectively; $N_{train}$ is the number of training scenes; and $N_{view}^i$ is the number of camera views of the $i$-th scenes.
The test data $\mathcal{D}_{test}$ is defined in a similar way.
During training, the generalizable NeRF model $\mathcal{M}(\cdot)$ renders novel views by aggregating the information of nearby source views of the same scene
\begin{equation}
    \hat{\mathbf{I}}_{r} = \mathcal{M}(\{\mathbf{I}_i, \mathbf{K}_i, \mathbf{E}_i\}_{i=1}^{N_{near}}, \mathbf{K}_{r}, \mathbf{E}_{r}),
    \label{eq:1}
\end{equation}
where $\mathbf{K}_{r}$ and $\mathbf{E}_{r}$ are the camera intrinsic and extrinsic of rendering view respectively; and $N_{near}$ is the number of source views.
Then we can train $\mathcal{M}(\cdot)$ by minimizing the loss between rendered images and ground truth
\begin{equation}
    \mathcal{M} = \mathop{\text{argmin}}_{\theta}\| \hat{\mathbf{I}}_{r} - \mathbf{I}_r \|^2_2.
\end{equation}
After training, we can render arbitrary views for unseen scenes by Eq.(\ref{eq:1}) without per-scene optimization.

\subsection{Synthetic-to-real Generalization}
\label{sec:Syn2Real}

In this paper, we consider the synthetic-to-real generalization of NeRF, which aims to train a NeRF model on synthetic data only and generalize it to real data.
It is practical because synthetic data is usually easier to obtain.
However, existing works of NeRF generalization~\cite{yu2021pixelnerf, chen2021mvsnerf, johari2022geonerf} mainly use real data as the training set.
A few works~\cite{wang2021ibrnet, zhang2022nerfusion} tried to use a small portion of synthetic data together with real data in model training, without evaluating the possible negative effects of synthetic data.
In this section, we first evaluate the effects of synthetic data in NeRF training via extensive experiments.
Specifically, we choose 3D-FRONT\cite{fu20213d} as our synthetic training set, which is a large-scale repository of synthetic indoor scenes with 18797 rooms.
ScanNet~\cite{dai2017scannet} is used as the test set, which is also a dataset about indoor scenes.
See Sec.\ref{sec:dataset} for more experimental details.
We adopt IBRNet~\cite{wang2021ibrnet} as the baseline considering its ease of use and promising performance.
For comparison, we also train the model on ScanNet.

As illustrated in Fig.\ref{fig:pred-sacnnet} and Fig.\ref{fig:pred-3dfront}, we observe that the model trained on synthetic data results in severer artifacts while better fine-grained details compared to the one trained on real data.
We further visualize the volume density along the ray for pixels with severe artifacts, as shown in Fig.\ref{fig:den-scannet} and Fig.\ref{fig:den-3dfront}.
We can see that the model trained on the synthetic data tends to predict volume densities with a sharper but less accurate distribution, while the model trained on the real data tends to be more conservative.
This is further demonstrated by Fig.\ref{fig:depth}.
The reason for these observations may be that the synthetic data is less noisy (real data usually involve inaccurate camera pose, image motion blur, and lighting changes), causing the model to be more confident in predictions and thus generate sharper densities.
However, when generalizing to real data which are noisy, the model may fail to accurately predict the geometry of the scene, resulting in serious artifacts in rendered images.
These observations inspire us to find a geometric-aware generalization method to solve the above-mentioned problem.


\section{Method}
\label{sec:method}

To tackle the above problem, we propose \textbf{ContraNeRF}, a generalizable NeRF method that combines contrastive learning with geometry information, enabling generalization from synthetic data to real data.
The overall framework is presented in Fig.\ref{fig:pipeline}, and the following sections provide details of our method.

\subsection{Geometry Aware Feature Extraction}
\label{sec:crossgam}

Given nearby source views $\{\mathbf{I}_i, \mathbf{K}_i, \mathbf{E}_i\}_{i=1}^{N_{near}}$, we first use a shared CNN to extract features $\mathbf{F}_i$ from each image $\mathbf{I}_i$.
Then we get the geometrically enhanced features $\{\mathbf{F}_i'\}_{i=1}^{N_{near}}$ by exchanging information between source views as described below.

Let $\mathbf{u}_i = [u, v]^{\top}$ denote the 2D coordinate of points on the $i$-th source view.
Firstly, we obtain the ray of point $\mathbf{u}_i$ as a line $\mathcal{R}$ in world coordinates parametrized by $\delta$ as 
\begin{equation}
    \mathcal{R}(\delta) = \mathbf{t}_i + \delta \mathbf{R}_i \mathbf{K}_i^{-1} [\mathbf{u}_i^{\top}, 1]^{\top},
    \label{eq:ray}
\end{equation}
where $\mathbf{K}_i$, $[\mathbf{R}_i, \mathbf{t}_i]$ represent the camera intrinsic and extrinsic parameters of the $i$-th source view respectively.
Then, we sample a sequence of points $\mathbf{p}^{s} = \mathcal{R}(\delta^s)$ along the ray $\mathcal{R}$ and project them to the $j$-th source view as follows 
\begin{equation}
    d_{j}^s[\mathbf{v}_{j}^{s \top}, 1]^\top = \mathbf{K}_j \mathbf{R}_j^{-1} (\mathbf{p}^{s} - \mathbf{t}_j), s=1,...,N_s,
    \label{eq:project}
\end{equation}
where $\mathbf{v}_j^s$ is the 2D coordinates of the projection in the $j$-th source view and $d_j^s$ is the corresponding depth; and $N_s$ is the number of sample points.
Now we have a collection of projections $\{\mathbf{v}_j^s\}_{s=1}^{N_{s}}$ for the $j$-th source view.

Then we enhance the feature $\mathbf{F}_i$ by aggregating the features from other source views via cross-view attention, as illustrated in Fig.\ref{fig:pipeline}.
There are two stages in the aggregation.
The first stage tries to aggregate the features of the projection points in the $j$-th source view.
Formally, let $\mathbf{f}_i$ denote the local feature of $\mathbf{F}_i$ at position $\mathbf{u}_i$ and $\mathbf{g}_j^s$ denote the local feature of $\mathbf{F}_j$ at position $\mathbf{v}_j^s$.
Then, we aggregate features $\{\mathbf{g}_j^s\}_{s=1}^{N_s}$ through Multi-Head Attention (MHA) layers
\begin{equation}
    \mathbf{g}_j = \textbf{MHA}(\mathbf{f}_i, \{\mathbf{g}_j^s + \mathcal{P}(s)\}_{s=1}^{N_s}),
\end{equation}
where $\mathbf{f}_i$ is the query of MHA; $\{\mathbf{g}_j^s + \mathcal{P}(s)\}_{s=1}^{N_s}$ serve as the keys and values of MHA; and $\mathcal{P}(s)$ is the position embedding of $s$.
Then, the second stage aims to aggregate the features of each source view and we achieve this through MHA too
\begin{equation}
    \mathbf{f}_i' = \textbf{MHA}(\mathbf{f}_i, \{ \mathbf{g}_j \}_{j=1, j\neq i}^{N_{near}}),
\end{equation}
where $\mathbf{f}_i$ is the query and $\{ \mathbf{g}_j \}_{j=1, j\neq i}^{N_{near}}$ are the keys and values of MHA.
We apply the above process for features of each source view and finally we get the geometrically enhanced features $\{\mathbf{F}_i'\}_{i=1}^{N_{near}}$.

Intuitively, our method tries to find the most similar features of $\mathbf{f}_i$ from other source views and aggregate them with $\mathbf{f}_i$.
As a result, similar features will become more similar, which is loosely similar to clustering.

\subsection{Geometry Aware Contrastive Learning}
\label{sec:geocontrast}

To better predict geometry, we enhance multi-view consistency through contrastive learning while taking into account the geometric constraints between views.
Here we describe how our \textbf{Geometry Aware Contrastive Learning} (\textbf{GeoContrast}) works in detail.

As illustrated in Fig.\ref{fig:pipeline}, we conduct contrastive learning between each pair of source views.
Take the $i$-th source view and the $j$-th source view as an example.
We first randomly sample a batch of pixels in the $i$-th source view, denoted as the $\{ \mathbf{p} \}_{N_c}$, where $\mathbf{p}=(p, q)$ is the 2D coordinate of the sampled pixel and $N_{c}$ is the number of samples.
For each sampled pixel $\mathbf{p}$ in the $i$-th source view, we specify the corresponding positive sample $\mathbf{q}_+$ and negative samples $\{ \mathbf{q_-} \}_{N_{neg}}$ in the $j$-th view according to ground truth depth, where $N_{neg}$ is the number of the negative samples.

\noindent \textbf{Positive pair.}
We take the pixels projected from the same 3D surface point in each source view as a positive pair.
Specifically, for a sample $\mathbf{p}$ of the $i$-th source view, we obtain the 3D point $\mathbf{p}_3$ that $\mathbf{p}$ is projected from as $\mathbf{p}_3=t_i + \delta_{\mathbf{p}} R_i K_i^{-1} \overline{\mathbf{p}}$ by taking Eq.(\ref{eq:ray}), where $\overline{\mathbf{p}}$ is the homogeneous coordinates of $\mathbf{p}$ and $\delta_{\mathbf{p}}$ is the depth.
Then we project 3D point $\mathbf{p}_3$ to the $j$-th source view as $\mathbf{q}_+$ by taking Eq.(\ref{eq:project}).
As there may be occlusions in the scene, we only consider $(\mathbf{p}, \mathbf{q}_+)$ of unoccluded regions as positive pairs.

\noindent \textbf{Negative pairs.}
We first project $\mathbf{p}$ to the $j$-th source view to get a collection of projections $\{ \mathbf{q}_- \}_{N_{neg}}$ by taking Eq.(\ref{eq:ray}) and Eq.(\ref{eq:project}).
Then we take $\{ (\mathbf{p}, \mathbf{q}_-) \}_{N_{neg}}$ as the negative pairs.
Here, each negative pair is the projection of 3D point under different source views, so this sampling strategy can help the model better capture geometric information.

After determining the positive and negative samples, the contrastive loss for $\mathbf{p}$ is defined as
\begin{equation}
    \mathcal{L}_{\mathbf{p}}^{\text{NCE}} = -\log \frac{\exp(\mathbf{p}' \cdot \mathbf{q}_+' / \tau)}{\exp(\mathbf{p}' \cdot \mathbf{q}_+' / \tau) + \sum_{\mathbf{q}_-}\lambda_{\mathbf{q}_-}\exp(\mathbf{p}' \cdot \mathbf{q}_-' / \tau)}.
    \label{eq:weight-cl}
\end{equation}
where $\mathbf{p}'$ is the local feature of $\mathbf{F}_i'$ at position $\mathbf{p}$;
$\mathbf{q}_+'$ and $\mathbf{q}_-'$ are the local features of $\mathbf{F}_j'$ at position $\mathbf{q}_+$ and $\mathbf{q}_-$ respectively;
$\mathbf{F}'_i$ is the output of cross-view attention for the $i$-th source view as mentioned in Sec.\ref{sec:crossgam}; 
$\tau$ is a learnable scalar temperature parameter; 
$\lambda_{\mathbf{q}_-}$ is the weight assigned to each $\mathbf{q}_-$.
We calculate the weight $\lambda_{\mathbf{q}_-}$ by considering the distance between $\mathbf{q}_+$ and $\mathbf{q_-}$
\begin{equation}
    \lambda_{\mathbf{q}_-} = N_{neg} \frac{\exp(\| \mathbf{q}_+ - \mathbf{q}_- \|_2 / \tau')}{\sum_{\mathbf{q}_-} \exp(\| \mathbf{q}_+ - \mathbf{q}_- \|_2 / \tau')},
\end{equation}
where $\tau'$ is a scalar temperature hyper-parameter, set by default to 10000.
This weight measures the similarity between the positive sample and the negative sample.
In this way, we can down-weight the influence of the negative samples that are similar to the positive sample.
Finally, we average the loss for all samples $\mathbf{p}$ of each source view pair to form the final contrastive loss $\mathcal{L}_{\text{contrast}}$:
\begin{equation}
    \mathcal{L}_{\text{contrast}} = \frac{1}{N_c}\sum_{\mathbf{p}}\mathcal{L}_{\mathbf{p}}^{\text{NCE}}.
\end{equation}

Another way to choose the negative samples of $\mathbf{p}$ is to sample pixels from the $j$-th source view randomly, which are commonly used in previous pixel-level contrastive learning methods \cite{xie2021propagate, wang2021dense, hu2021region}.
However, this sampling strategy does not consider the geometric constraints in the 3D scenarios.
Intuitively, in our method, all the negative pairs are the projections of the non-surface 3D points and the positive pairs are the projections of the surface 3D points.
It will be easier to model scene geometry by the following network since our GeoContrast makes the non-surface points and surface points more distinguishable, and experiments show that our sampling strategy performs much better (see Sec.\ref{sec:ablation}).

\subsection{Rendering and Training}
\label{sec:render}

Following IBRNet\cite{wang2021ibrnet}, we calculate colors and densities for 3D points along the rendering ray $\mathbf{r}$ by checking the consistency among the features of each source view.
During rendering, different from the volume rendering equation that is commonly used in most previous NeRF-related works\cite{mildenhall2021nerf, wang2021ibrnet, chen2021mvsnerf, yu2021pixelnerf}, we accumulate colors along the ray weighted by densities after softmax.
\begin{equation}
    \hat{C}(\mathbf{r}) = \frac{1}{\sum_i\exp(\sigma_i)}\sum_i c_i \cdot \exp(\sigma_i),
    \label{eq:render}
\end{equation}
where $c_i$, $\sigma_i$ are the color and density for the $i$-th 3D sample point on the ray $\mathbf{r}$ respectively.
We find that Eq.(\ref{eq:render}) works well in our experiments and it can speed up the convergence of network training without affecting model performance, which is also explored in \cite{shi2021self}.
It may be because the ray transformer in IBRNet\cite{wang2021ibrnet} already has the ability to simulate light transport and occlusion in the radiance field.

\begin{figure*} 
    \centering
    \begin{subfigure}{0.165\linewidth}
        \includegraphics[width=\textwidth]{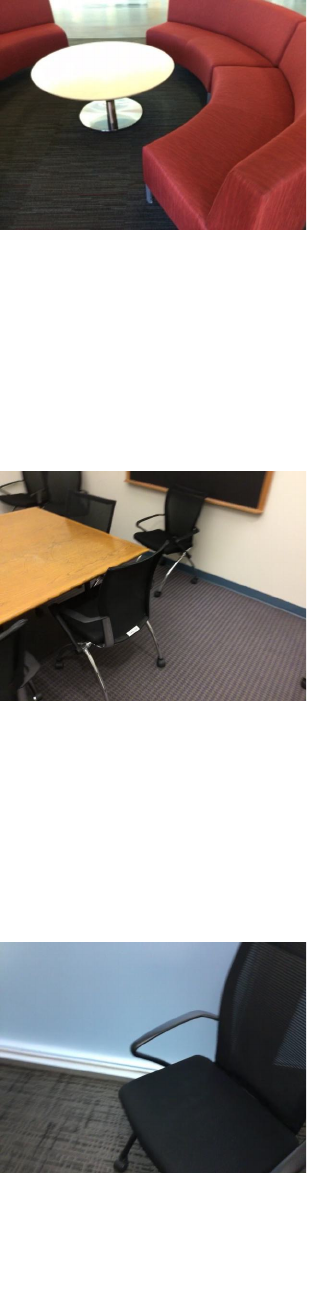}
        \caption{Ground Truth}
        \label{fig:gt}
    \end{subfigure}
    \hspace{-2mm}
    \begin{subfigure}{0.165\linewidth}
        \includegraphics[width=\textwidth]{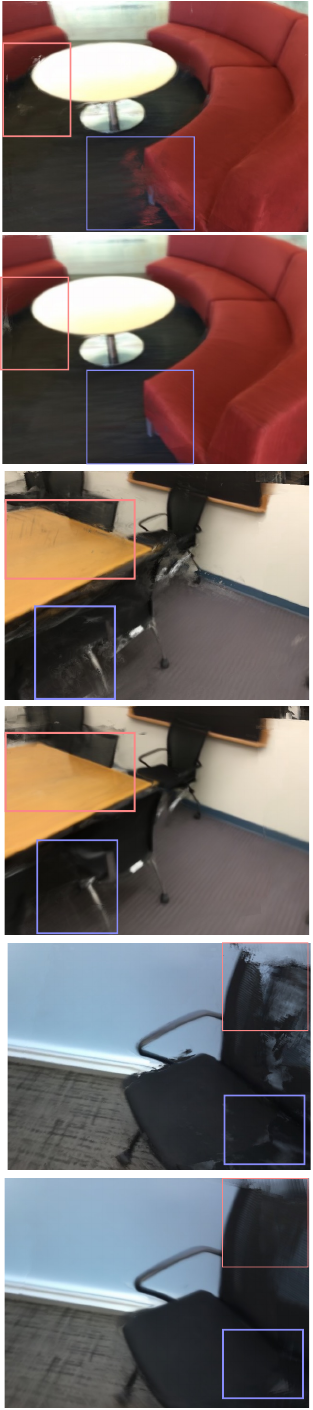}
        \caption{IBRNet}
        \label{fig:ibrnet}
    \end{subfigure}
    \hspace{-2mm}
    \begin{subfigure}{0.165\linewidth}
        \includegraphics[width=\textwidth]{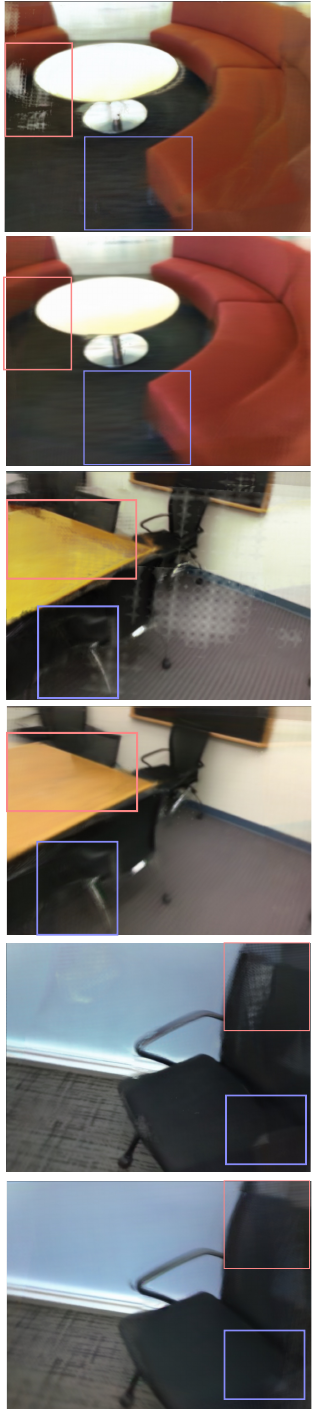}
        \caption{MVSNeRF}
        \label{fig:mvsnerf}
    \end{subfigure}
    \hspace{-1.5mm}
    \begin{subfigure}{0.165\linewidth}
        \includegraphics[width=\textwidth]{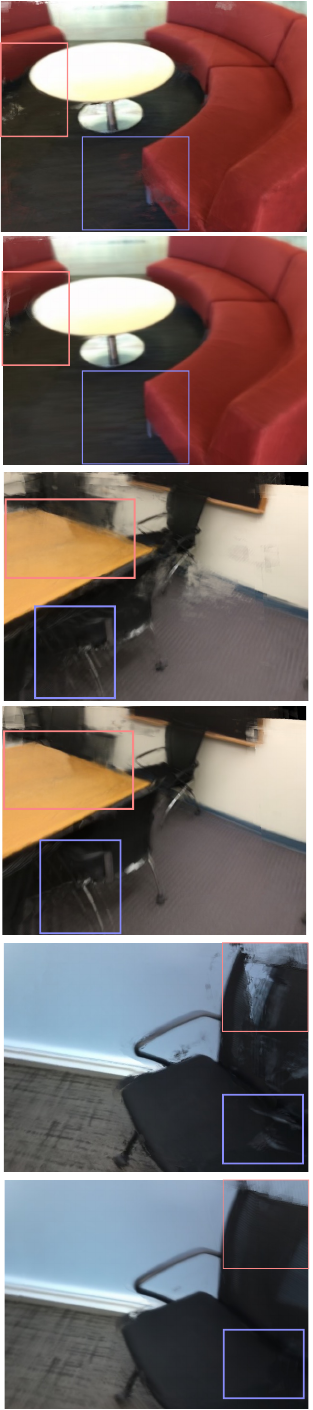}
        \caption{GeoNeRF}
        \label{fig:geonerf}
    \end{subfigure}
    \hspace{-3mm}
    \begin{subfigure}{0.167\linewidth}
        \includegraphics[width=\textwidth]{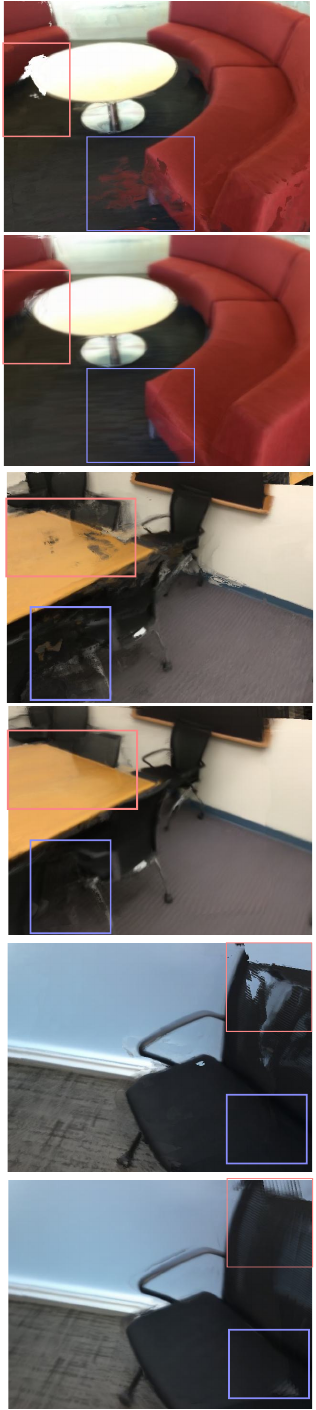}
        \caption{Neuray}
        \label{fig:neuray}
    \end{subfigure}
    \hspace{-2mm}
    \begin{subfigure}{0.1836\linewidth}
        \includegraphics[width=\textwidth]{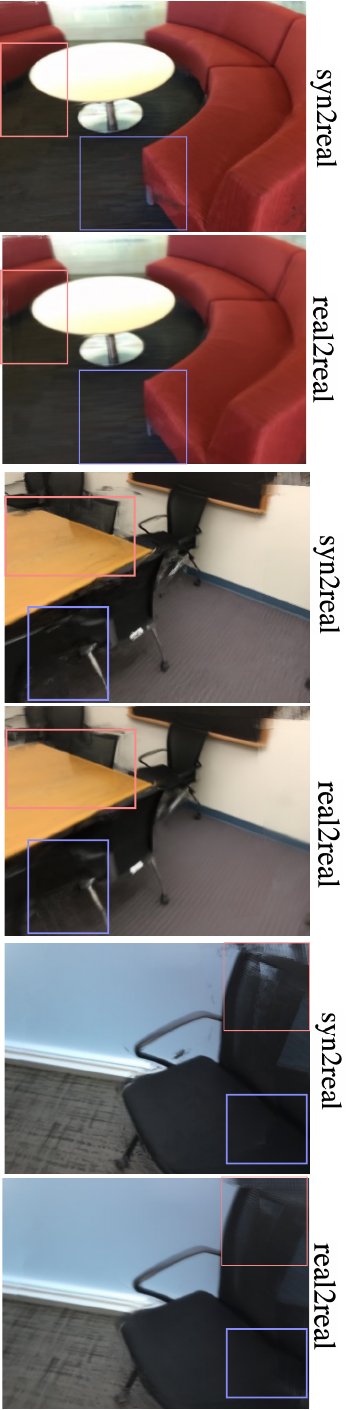}
        \caption{Ours}
        \label{fig:ours}
    \end{subfigure}
    \caption{\textbf{Qualitative comparison on ScanNet dataset\cite{dai2017scannet}.} For each scene, two kinds of results for each method are shown, where the first row is the result after training on synthetic data and the second row is the result after training on real data. Our model more accurately preserves the details while it generates fewer artifacts than other generalizable NeRF methods when training on synthetic data.}
    \label{fig:vis-scannet}
\end{figure*}

Following \cite{wang2021ibrnet, liu2022neural}, we use the coarse-to-fine sampling strategy with 64 sample points in both stages.
Then we can get the color loss
\begin{equation}
    \mathcal{L}_{\text{color}} = \sum_{\mathbf{r}\in\mathcal{R}}\Bigg[\Big\|\hat{C}_c(\mathbf{r}) - C(\mathbf{r})\Big\|_2^2 + \Big\|\hat{C}_f(\mathbf{r}) - C(\mathbf{r})\Big\|_2^2 \Bigg],
\end{equation}
where $\mathcal{R}$ is the set of rays in each batch; and $\hat{C}_c(\mathbf{r})$, $\hat{C}_f(\mathbf{r})$, and $C(\mathbf{r})$ are the coarse stage RGB prediction, fine stage RGB prediction, and ground truth for ray $\mathbf{r}$ respectively.
Our final loss function is the sum of the contrastive loss and color loss.
\begin{equation}
    \mathcal{L}_{\text{total}} = \mathcal{L}_{\text{contrast}} + \mathcal{L}_{\text{color}}.
\end{equation}

\section{Experiments}
\label{sec:experiments}

\subsection{Experimental Settings}
\label{sec:dataset}

\noindent \textbf{Datasets.} In synthetic-to-real generalization, we choose 3D-FRONT\cite{fu20213d} and ScanNet\cite{dai2017scannet} as our synthetic training set and real test set respectively. (1) \textbf{3D-FRONT} is a large-scale, and comprehensive repository of synthetic indoor scenes.
It contains 18,797 rooms diversely furnished by 3D objects.
Following the data partition strategy in \cite{yu2021pixelnerf, chen2021mvsnerf}, we randomly sample 88 scenes from 3D-FRONT as our synthetic training datasets.
For each scene in 3D-FRONT, we sample 200 camera views and render each view at 640 × 480 resolution using BlenderProc\cite{denninger2019blenderproc}.
(2) \textbf{ScanNet} is an RGB-D video dataset containing more than 1500 scans with 2.5 million views. 
Each scene contains 1K–5K views. 
We uniformly sample one-tenth of views and resize each image to a resolution of 640 × 480 for use.
In our experiments, we randomly select 88 scenes of ScanNet as our real training datasets and 8 scenes of ScanNet as our test datasets.
On each test scene, we leave out 1/8 number of images as test views and the rest images as source views following\cite{wang2021ibrnet, liu2022neural, johari2022geonerf}.

\noindent \textbf{Baselines and evaluation metrics.} We compare our method with state-of-the-art generalizable NeRF methods, including PixelNeRF\cite{yu2021pixelnerf}, IBRNet\cite{wang2021ibrnet}, MVSNeRF\cite{chen2021mvsnerf}, GeoNeRF\cite{johari2022geonerf} and Neuray\cite{liu2022neural}.
Following IBRNet,
we evaluate all these methods using PSNR, SSIM, and LPIPS.

\noindent  \textbf{Implementation details.} 
In contrastive learning, we sample a batch of 576 pixels for training and sample 512 negative pairs for each positive pair, where the parameters are tuned.
Coarse and fine models share the same feature extractor.
To render a novel view, we use 10 neighboring input views as the source views, and we randomly sample 512 pixels from the novel view as a batch during training following\cite{wang2021ibrnet}.
We train the whole pipeline for 100k iterations using Adam optimizer\cite{kingma2014adam} and the base learning rate is $10^{-3}$.
To achieve fair and accurate comparisons, we run all methods on the same experiment settings and use the official code to run the experiments.
All experiments are conducted on the V100 GPU.

\subsection{Results}

\begin{table}
  \centering
  \scalebox{0.85}{
  \begin{tabular}{cccc}
    \toprule
    Method & PSNR $\uparrow$ & SSIM $\uparrow$ & LPIPS $\downarrow$ \\
    \midrule
    PixelNeRF\cite{yu2021pixelnerf} & 20.19 (22.44) & 0.736 (0.774) & 0.511 (0.450) \\
    IBRNet\cite{wang2021ibrnet} & 23.67 (25.25) & 0.807 (0.840) & 0.355 (0.328)\\
    MVSNeRF\cite{chen2021mvsnerf} & 22.90 (24.90) & 0.793 (0.824) & 0.408 (0.357) \\
    GeoNeRF\cite{johari2022geonerf} & 23.67 (25.18) & 0.797 (0.837) & 0.349 (0.327) \\
    Neuray\cite{liu2022neural} & 22.75 (25.22) & 0.785 (0.838) & 0.369 (0.325) \\
    Ours & \textbf{24.81} (\textbf{25.58}) & \textbf{0.831} (\textbf{0.847}) & \textbf{0.333} (\textbf{0.320}) \\
    \bottomrule
  \end{tabular}
  }
  \caption{\textbf{Quantitative comparisons on the ScanNet dataset\cite{dai2017scannet}}. All methods are trained on the same scenes and tested on unseen real scenes. We report PSNR/SSIM (higher is better) and LPIPS (lower is better). The results of the model trained with synthetic data are shown outside the brackets, and the results of the model trained with real data are shown in brackets. Our method quantitatively outperforms prior work on all metrics.}
  \label{tab:scannet}
  \vspace{-5mm}
\end{table}

We show the quantitative results in Tab.\ref{tab:scannet} and visual comparisons in Fig.\ref{fig:vis-scannet}. Tab.\ref{tab:scannet} shows the superiority of our method with respect to the previous generalizable NeRF models on the synthetic-to-real setting.
Note that some previous works, such as MVSNeRF\cite{chen2021mvsnerf} and GeoNeRF\cite{johari2022geonerf}, also attempt to incorporate geometry into the model for performance improvement, but our approach differs from them in two ways.
On the one hand, when performing multi-view feature fusion, our method samples 3D points randomly with an inverse depth distribution, while MVSNeRF and GeoNeRF only sample points at preset discrete positions to build the cost volume, which may introduce quantization errors.
Moreover, with the help of contrastive learning, our method introduces consistency among multi-view features in a more straightforward way, while MVSNeRF and GeoNeRF only fuse the multi-view features without explicitly considering the consistency.
These two points allow us to achieve better results.
As shown in Fig.\ref{fig:vis-scannet}, our method can produce images with fine-grained details in both geometry and appearance, and it generates fewer artifacts compared with the previous generalizable NeRF methods under both the synthetic-to-real and real-to-real settings.
Note that in synthetic-to-real case, the interpolation-based generalizable NeRF method~\cite{wang2021ibrnet, johari2022geonerf, liu2022neural} performs better on color prediction than the method using the network to predict color~\cite{chen2021mvsnerf}.
This is because the color predicted by the interpolation-based methods comes from the input images, so there is no domain gap even under the synthetic-to-real setting.
Meanwhile, we can see that the model trained on synthetic data can retain more detail than the model trained on real data.

\subsection{Other Benchmark Datasets}
\label{sec: other-data}

\begin{table}
  \centering
  \scalebox{0.85}{\begin{tabular}{cccc}
    \toprule
    Method & PSNR $\uparrow$ & SSIM $\uparrow$ & LPIPS $\downarrow$ \\
    \midrule
    PixelNeRF\cite{yu2021pixelnerf} & 19.40 & 0.463 & 0.447 \\
    IBRNet\cite{wang2021ibrnet} & 25.76 & 0.861 & 0.173 \\
    MVSNeRF\cite{chen2021mvsnerf} & 23.83 & 0.723 & 0.286 \\
    GeoNeRF\cite{johari2022geonerf} & 26.49 & 0.883 & 0.153 \\
    Neuray\cite{liu2022neural} & 26.47 & 0.875 & 0.158 \\
    Ours & \textbf{27.69} & \textbf{0.904} & \textbf{0.129} \\
    \bottomrule
  \end{tabular}}
  \caption{\textbf{Quantitative comparisons on the DTU dataset\cite{jensen2014large}}. Our model is able to generate better results than previous state-of-the-art generalization NeRF models.}
  \label{tab:dtu}
  \vspace{-1mm}
\end{table}

\begin{table}
  \centering
  \scalebox{0.85}{\begin{tabular}{cccc}
    \toprule
    Method & PSNR $\uparrow$ & SSIM $\uparrow$ & LPIPS $\downarrow$ \\
    \midrule
    PixelNeRF\cite{yu2021pixelnerf} & 18.66 & 0.588 & 0.463 \\
    IBRNet\cite{wang2021ibrnet} & 25.17 & 0.813 & 0.200 \\
    MVSNeRF\cite{chen2021mvsnerf} & 21.18 & 0.691 & 0.301 \\
    GeoNeRF\cite{johari2022geonerf} & \textbf{25.44} & 0.839 & 0.180 \\
    Neuray\cite{liu2022neural} & 25.35 & 0.818 & 0.198 \\
    Ours & \textbf{25.44} & \textbf{0.842} & \textbf{0.178}\\
    \bottomrule
  \end{tabular}}
  \caption{\textbf{Quantitative comparisons on the LLFF dataset\cite{mildenhall2019local}}. Our model is able to generate better results than previous state-of-the-art generalization NeRF models.}
  \label{tab:llff}
  \vspace{-1mm}
\end{table}

\begin{table}
  \centering
  \scalebox{1}{
  \scalebox{0.73}{\begin{tabular}{cccc}
    \toprule
     & PSNR $\uparrow$ & SSIM $\uparrow$ & LPIPS $\downarrow$ \\
    \midrule
    BaseModel & 23.71 (25.27) & 0.810 (0.840) & 0.352 (0.327) \\
    Random negative sampling & 23.84 (25.30) & 0.814 (0.840) & 0.347 (0.326)\\
    GeoContrast(w/o weight) & 24.28 (25.40) & 0.821 (0.843) & 0.339 (0.325) \\
    GeoContrast & 24.53 (25.45) & 0.825 (0.843) & 0.337 (0.324) \\
    Cross-view attention & 24.25 (25.47) & 0.820 (0.844) & 0.342 (0.322) \\
    Full model Ours & \textbf{24.81} (\textbf{25.58}) & \textbf{0.831} (\textbf{0.847}) & \textbf{0.333} (\textbf{0.320}) \\
    \bottomrule
  \end{tabular}}
  }
  \caption{\textbf{Ablation study on the ScanNet dataset\cite{dai2017scannet}}. The results of synthetic/real data are shown outside/in brackets, respectively. Refer to Sec.\ref{sec:ablation} for details.}
  \label{tab:ablation}
\end{table}

To further demonstrate the effectiveness of our method, we also conduct experiments in the settings described in the previous generalizable NeRF methods~\cite{yu2021pixelnerf,wang2021ibrnet,chen2021mvsnerf,johari2022geonerf,liu2022neural}.

\noindent \textbf{Evaluation datasets.}
We consider two widely adopted benchmarks, including the DTU dataset\cite{jensen2014large} and LLFF dataset\cite{mildenhall2019local}.
Following \cite{liu2022neural}, we select four objects (birds, tools, bricks, and snowman) as test objects for DTU dataset, and the test images of DTU dataset all use black backgrounds.
On each test scene, we use 1/8 images as test views and the evaluation resolution is 800 × 600 for the DTU dataset, and 1008 × 756 for the LLFF dataset.

\noindent \textbf{Training datasets.}
Following \cite{liu2022neural, wang2021ibrnet}, we use both the synthetic and real data for model training.
For synthetic data, we use the Google Scanned Object dataset\cite{downs2022google}. 
For real data, we use forward-facing training datasets\cite{mildenhall2019local, wang2021ibrnet} and the rest training objects from the DTU dataset.

\begin{figure}[H]
\vspace{-5mm}
    \centering
    \begin{subfigure}{0.3\linewidth}
        \includegraphics[width=\textwidth]{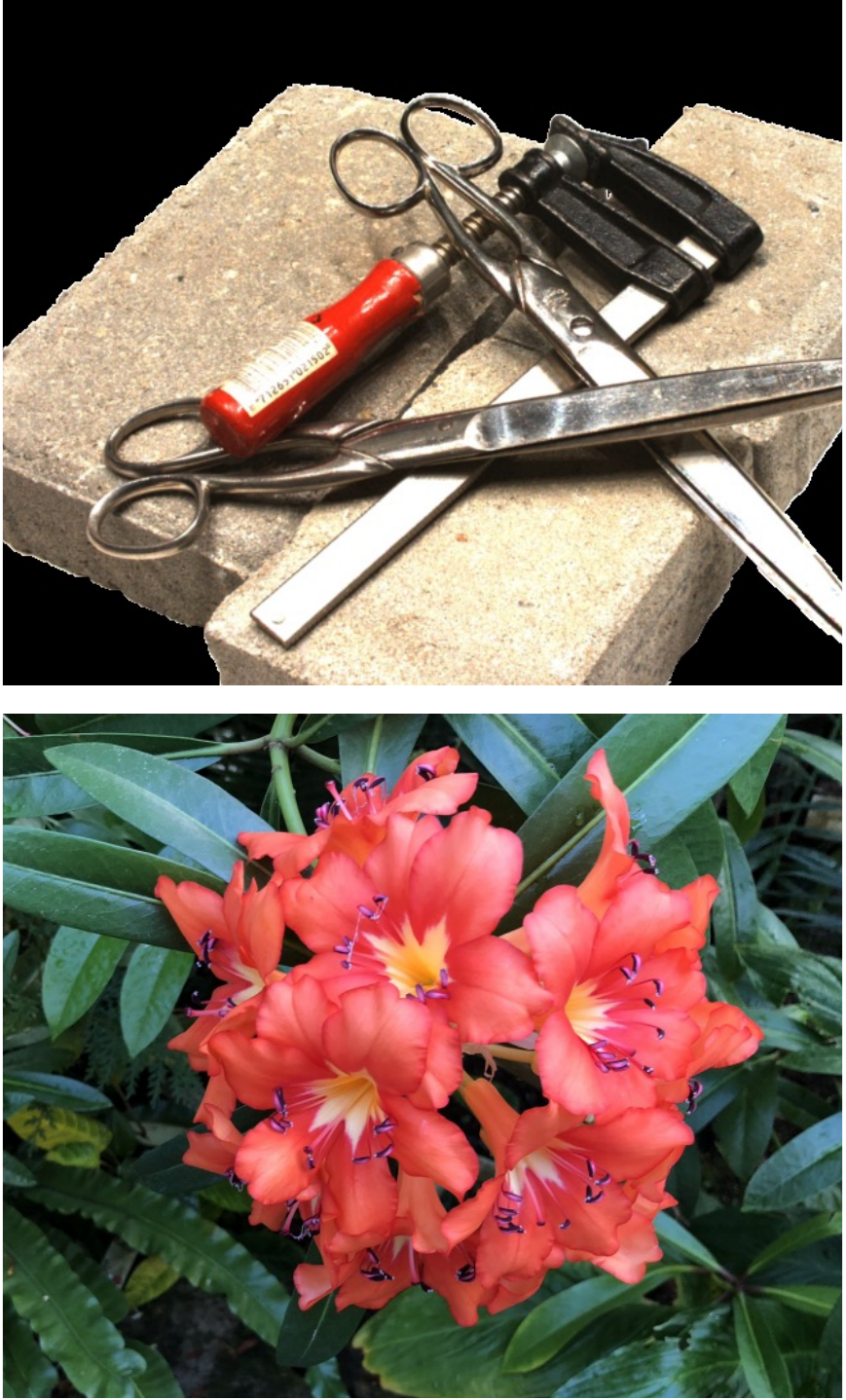}
        \caption{Ground Truth}
        \label{fig:visual-other-gt}
    \end{subfigure}
    \begin{subfigure}{0.3\linewidth}
        \includegraphics[width=\textwidth]{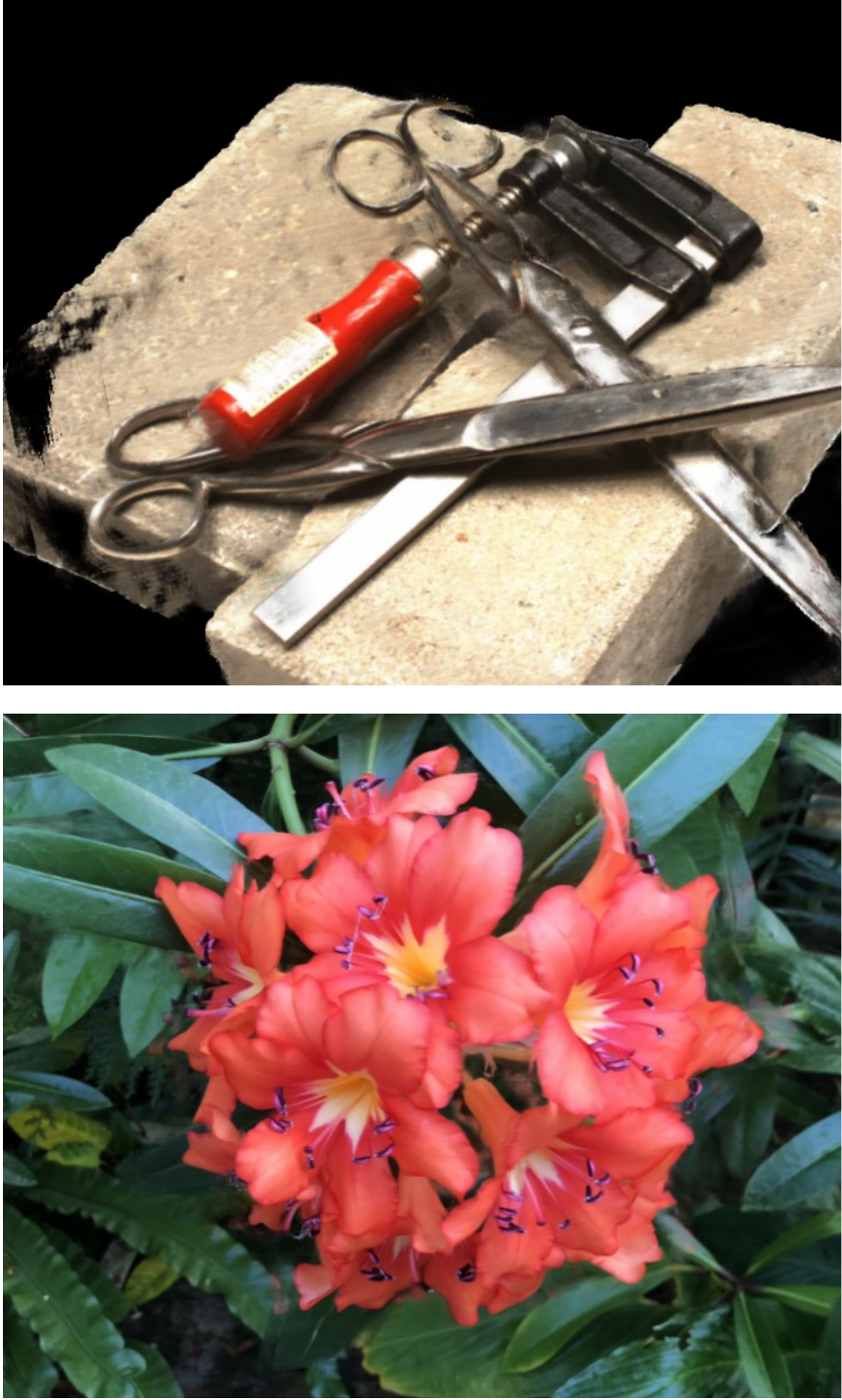}
        \caption{GeoNeRF}
        \label{fig:visual-other-geo}
    \end{subfigure}
    \begin{subfigure}{0.3\linewidth}
        \includegraphics[width=\textwidth]{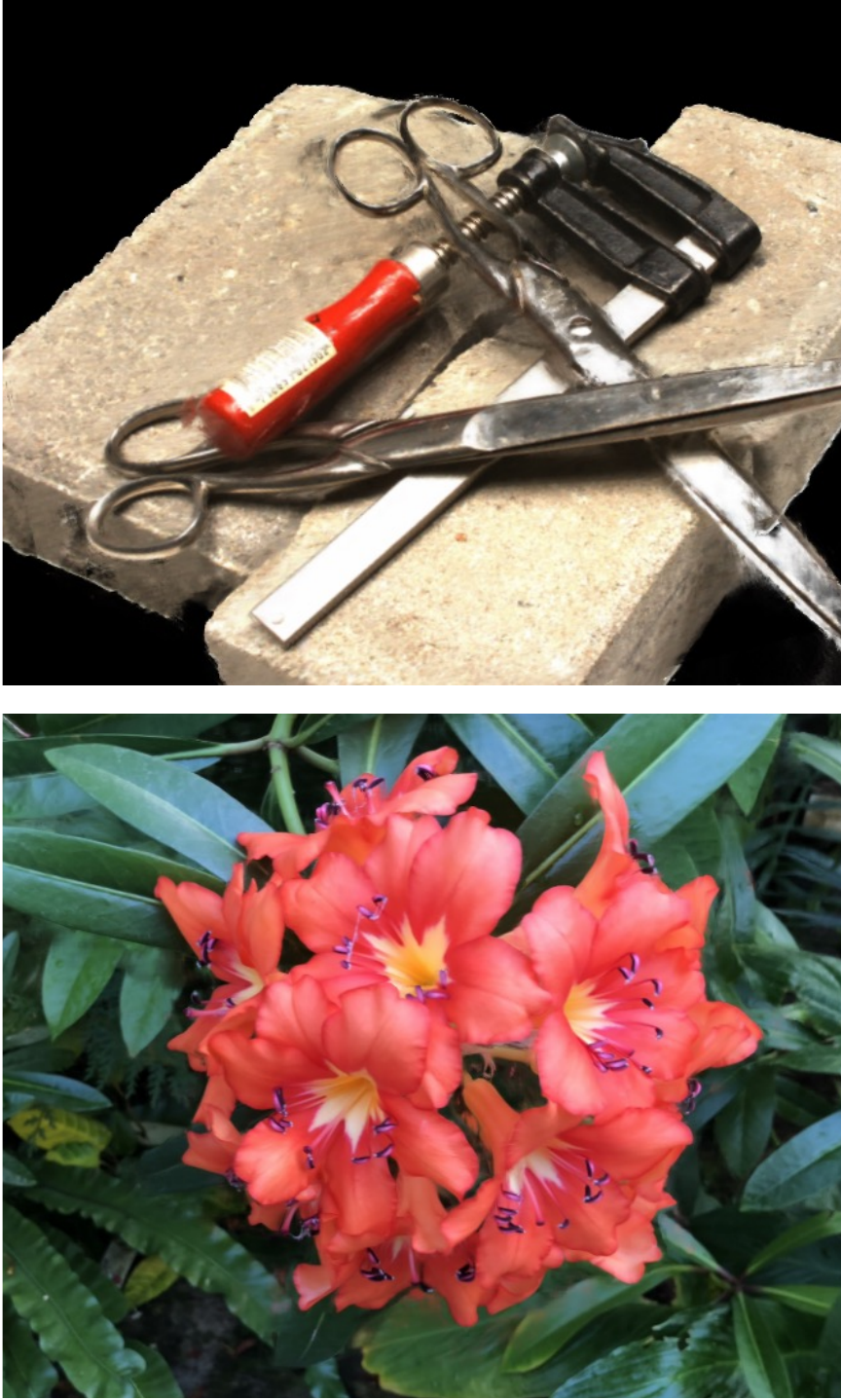}
        \caption{Ours}
        \label{fig:visual-other-our}
    \end{subfigure}
    \caption{\textbf{Qualitative comparison on DTU dataset\cite{jensen2014large} and LLFF dataset\cite{mildenhall2019local}.} Our method can render images with fewer artifacts. 
    }
    \label{fig:visual-other}
\end{figure}

\noindent \textbf{Results.}
The quantitative results of DTU dataset and LLFF dataset are shown in Tab.\ref{tab:dtu} and Tab.\ref{tab:llff} respectively.
Our method works well on both DTU dataset and LLFF dataset.
On DTU dataset, our method outperforms previous state-of-the-art generalizable NeRF methods by a large margin.
This shows that our method can still work well even with a mixture of synthetic and real data.
Fig.\ref{fig:visual-other} shows some visualization results on the DTU dataset and LLFF datasets.
Similar to the results on ScanNet dataset, our model can produce images that are perceptually more similar to the ground truth than other methods.

\subsection{Ablation study}
\label{sec:ablation}

\noindent \textbf{Ablation on network design.} We conduct ablation studies in the settings introduced in Sec.\ref{sec:dataset} to validate the effectiveness of different design decisions.
We first train a model, called `BaseModel', that does not use cross-view attention and GeoContrast.
Note that the `BaseModel' still outperforms IBRNet\cite{wang2021ibrnet} since we use Eq.(\ref{eq:render}) as the rendering equation.
For `Random negative sampling', we apply vanilla contrastive learning in `BaseModel', where negative
pairs are sampled from other views randomly, as mentioned in Sec.\ref{sec:geocontrast}.
It works, but the improvement is marginal.
Then we replace vanilla contrastive learning with our GeoContrast as shown at the third and fourth rows of Tab.\ref{tab:ablation}, where `GeoContrast(w/o weight)' means the GeoContrast without weighted contrastive loss in Eq.(\ref{eq:weight-cl}).
We can see that `GeoContrast(w/o weight)' can bring improvement compared to random sampling, thanks to the introduction of geometric constraints.
When equipped with weighted contrastive loss, GeoContrast can be further improved as shown at the 4-th rows of Tab.\ref{tab:ablation}.
The results of the 5-th rows show that our cross-view attention is also helpful for the generalization.
Finally, we combine cross-view attention with GeoContrast to form a complete model and it achieves the best results.

\begin{figure}
    \centering
    \includegraphics[width=0.9\linewidth]{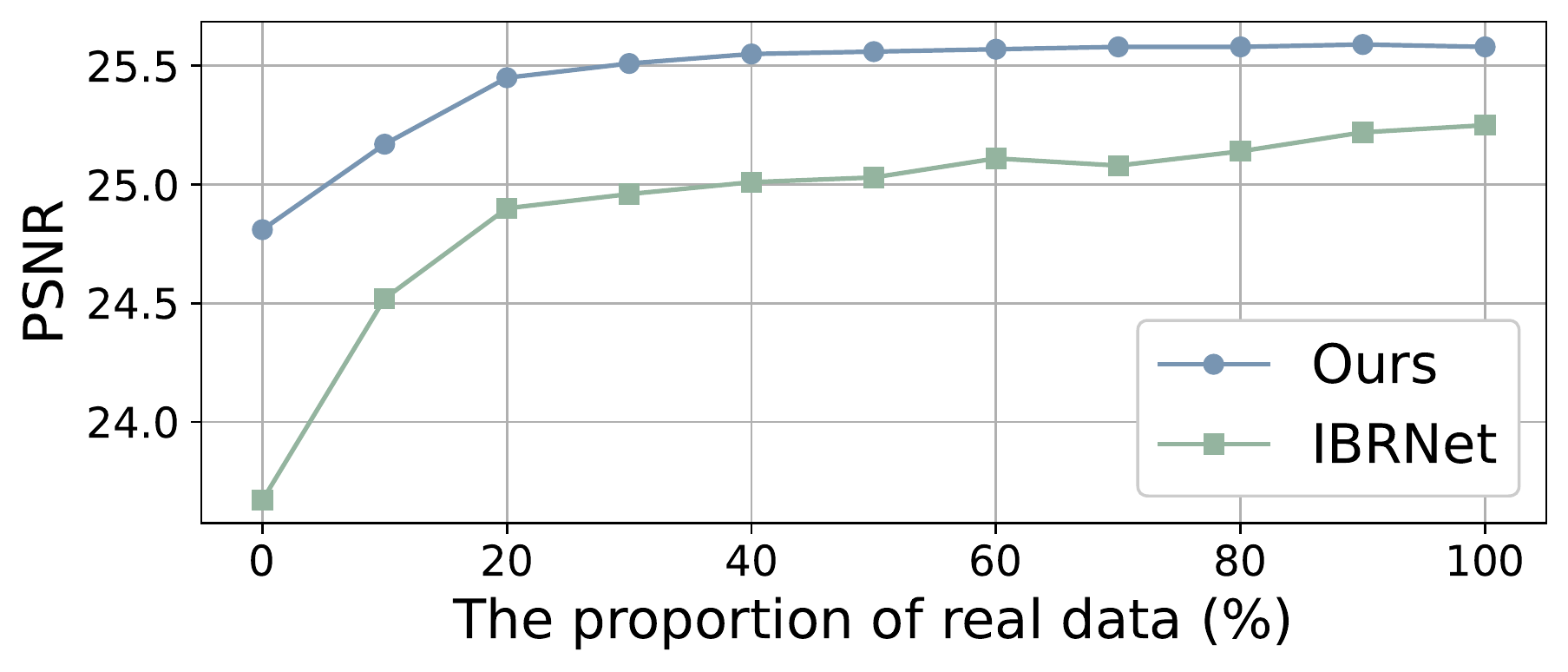}
    \caption{\textbf{Curves of PSNR of our method and IBRNet\cite{wang2021ibrnet} with different proportions of real and synthetic data.}}
    \label{fig:propotion}
\end{figure}

\noindent \textbf{Ablation on proportion of real data.} Although our method makes better use of synthetic data, there is still a performance gap between the model trained on synthetic data and the model trained on real data.
A natural question is can we use a mix of real and synthetic data to boost model performance and how much real data do we need?
Here, we conduct experiments with different proportions of real and synthetic data on our model and IBRNet\cite{wang2021ibrnet}.
Fig.\ref{fig:propotion} shows the PSNR under varying proportions of real and synthetic data.
We find that as the proportion of real data gradually increases, the performance of the model is improved.
However, when the proportion of real data reaches a certain value, the performance will not continue to improve.
For example, in our method, when the proportion of real data reaches 40\%, the performance of the model saturates.
This means that we only need to use a small amount of real data and a certain amount of synthetic data to achieve the same effect as using real data completely, while IBRNet needs to use more real data to achieve better results.

\section{Conclusion}
\label{sec:conclusion}

We present a generalizable neural radiance field method for synthetic-to-real novel view synthesis.
Unlike the real-to-real novel view synthesis, models trained on synthetic data tend to predict sharper but less accurate volume densities on real data, which may result in severe artifacts in rendered images.
To address this problem, we introduce geometry-aware contrastive learning to enable better modeling of scene geometry, thereby improving the model's ability to generalize from synthetic data to real data.
Experiments demonstrate that our method can render high-quality images while preserving fine details in the synthetic-to-real setting.

\paragraph{Acknowledgement.}
This work is supported by the National Key R\&D Program of China (2022ZD0114900), the National Science Foundation of China (NSFC62276005), the National Research Foundation, Singapore under its AI Singapore Programme (AISG Award
No: AISG2-RP-2021-024), and the Tier 2 grant MOE-T2EP20120-0011 from the Singapore Ministry of Education.
We gratefully acknowledge the support of MindSpore, CANN (Compute Architecture for Neural Networks) and Ascend AI Processor used for this research.

{\small
\bibliographystyle{ieee_fullname}
\bibliography{egbib}
}

\newpage

\appendix
\section*{Appendix}

\section{Details of Preliminary Experiments}
\label{app: preliminary}
\setcounter{page}{1}

To draw the histograms shown in Fig.\ref{fig:depth}, we need to calculate the deviation and error of the predicted depth for each pixel in the test dataset.
Here we choose 3D-FRONT\cite{fu20213d} as our synthetic training set and ScanNet\cite{dai2017scannet} as our real training set and test set.
Refer to Sec.\ref{sec:dataset} and App.\ref{app: 3dfront} for more details on the dataset.
We train the models on the 3D-FRONT and ScanNet respectively and use the trained models to predict depth on the test dataset.
For each pixel $\mathbf{u}$ in the test dataset, we first obtain the ray $\mathbf{r}$ as a line $\mathcal{R}$ as shown in Eq.\ref{eq:ray} and sample a sequence of points $\{\mathbf{p}^s = \mathcal{R}(\delta^s)\}_{s=1}^{N_s}\}$, where $\delta^s$ is the depth of $\mathbf{p}^s$ and $N_s$ is the number of sampling points, set by 128 default.
Then we calculate the volume density $\sigma^s$ at each sampling point, just like we calculate the volume density when rendering the image.
Note that we use the fine stage predictions if coarse-to-fine sampling is applied.
We use the following formula to assign a weight $w^s$ to each sampling point according to the volume density $\sigma^s$
\begin{equation}
    w^s = (1 - \exp(-\sigma^s)) \cdot \exp(-\sum_{t=1}^{s-1}\sigma^t).
\end{equation}
Then we predict the depth $\hat{D}(\mathbf{u})$, the standard deviation of the depth $S(\mathbf{u})$, and the depth error $E(\mathbf{u})$ for each pixel $\mathbf{u}$
\begin{equation}
\begin{aligned}
    \hat{D}(\mathbf{u}) &= \sum_{s=1}^{N_s} w^s \cdot \delta^s, \\
    S(\mathbf{u}) &= \bigg( \sum_{s=1}^{N_s} w^s \cdot (\delta^s - \hat{D}(\mathbf{u}))^2 \bigg)^{\frac{1}{2}}, \\
    E(\mathbf{u}) &= \| \hat{D}(\mathbf{u}) - D(\mathbf{u}) \|,
\end{aligned}
\end{equation}
where $D(\mathbf{u})$ is the ground truth depth for pixel $\mathbf{u}$.
Finally, we can draw the histogram shown in Fig.\ref{fig:depth} based on $S(\mathbf{u})$ and $E(\mathbf{u})$.

\begin{figure*}
  \centering
  \begin{subfigure}{0.24\linewidth}
    \includegraphics[width=\textwidth]{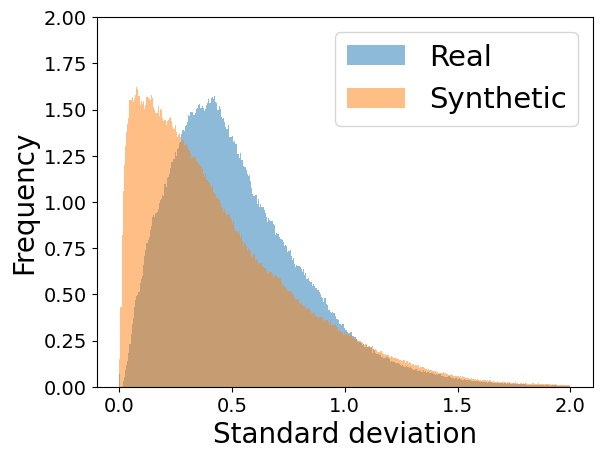}
    \caption{Deviation (MVSNeRF).}
    \label{fig: mvsnerf-var}
  \end{subfigure}
  \begin{subfigure}{0.24\linewidth}
    \includegraphics[width=\textwidth]{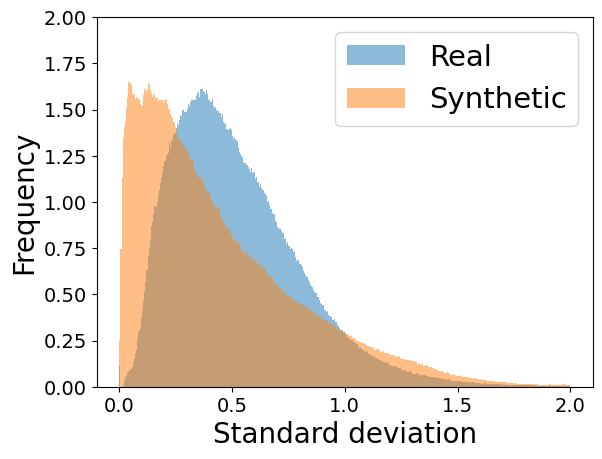}
    \caption{Deviation (GeoNeRF).}
    \label{fig: geonerf-var}
  \end{subfigure}
  \begin{subfigure}{0.24\linewidth}
    \includegraphics[width=\textwidth]{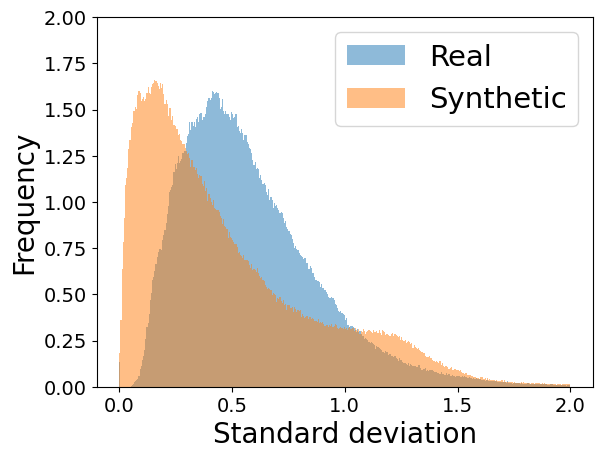}
    \caption{Deviation (Neuray).}
    \label{fig: neuray-var}
  \end{subfigure}
  \begin{subfigure}{0.24\linewidth}
    \includegraphics[width=\textwidth]{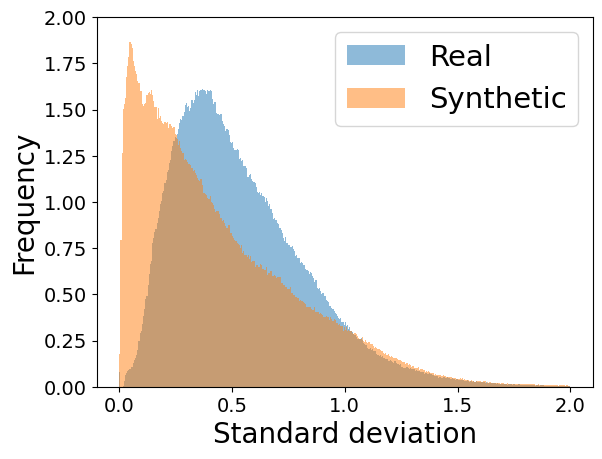}
    \caption{Deviation (Ours).}
    \label{fig: ours-var}
  \end{subfigure}
  \\
  \begin{subfigure}{0.24\linewidth}
    \includegraphics[width=\textwidth]{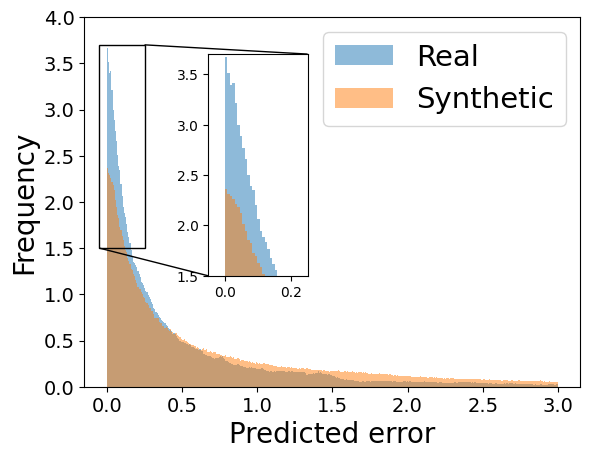}
    \caption{Error (MVSNeRF).}
    \label{fig: mvsnerf-err}
  \end{subfigure}
  \begin{subfigure}{0.24\linewidth}
    \includegraphics[width=\textwidth]{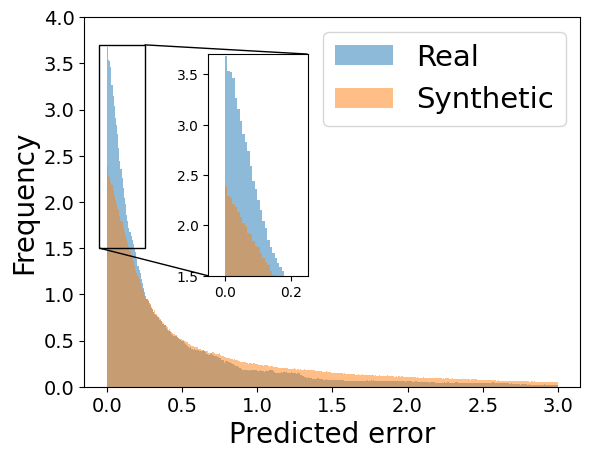}
    \caption{Error (GeoNeRF).}
    \label{fig: geonerf-err}
  \end{subfigure}
  \begin{subfigure}{0.24\linewidth}
    \includegraphics[width=\textwidth]{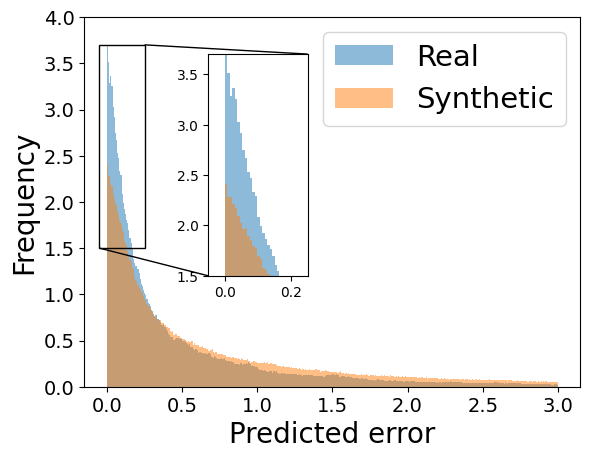}
    \caption{Error (Neuray).}
    \label{fig: neuray-err}
  \end{subfigure}
  \begin{subfigure}{0.24\linewidth}
    \includegraphics[width=\textwidth]{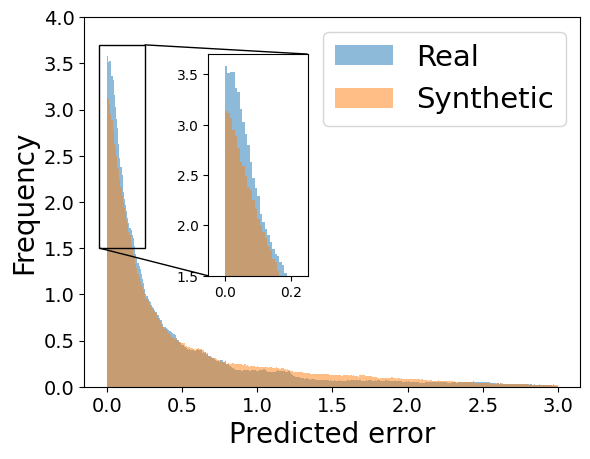}
    \caption{Error (Ours).}
    \label{fig: ours-err}
  \end{subfigure}
  \caption{\textbf{Deviation and error of predicted depth when trained
with synthetic and real data, respectively}. We plot the deviation and error of the predicted depth as the histogram for MVSNeRF\cite{chen2021mvsnerf} (column 1), GeoNeRF\cite{johari2022geonerf} (column 2), Neuray\cite{liu2022neural}, and our method (column 4). Our method is able to predict more accurate depth while maintaining its sharpness under synthetic-to-real setting.}
  \label{fig: appendix-depth}
\end{figure*}

\paragraph{Results of other methods}
We also calculate the deviation and error of the depth predicted by other generalizable NeRF models (MVSNeRF\cite{chen2021mvsnerf}, GeoNeRF\cite{johari2022geonerf}, Neuray\cite{liu2022neural}, and our method) as shown in Fig.\ref{fig: appendix-depth}.
Note that for our method, we calculate the weight $w^s$ following the Eq.\ref{eq:render} as 
\begin{equation}
    w^s = \frac{\exp(\sigma^s)}{\sum_{t=1}^{N_s}\exp(\sigma^t)}.
\end{equation}
Like IBRNet\cite{wang2021ibrnet}, previous NeRF generalization methods\cite{chen2021mvsnerf, johari2022geonerf, liu2022neural} tend to predict radiance fields that are sharper but less geometrically accurate under the synthetic-to-real setting.
In comparison, our method predicts a more accurate radiance field while remaining sharp.

\section{Preprocess of Dataset}
\label{app: 3dfront}

\begin{figure*}
    \centering
    \includegraphics[width=0.99\textwidth]{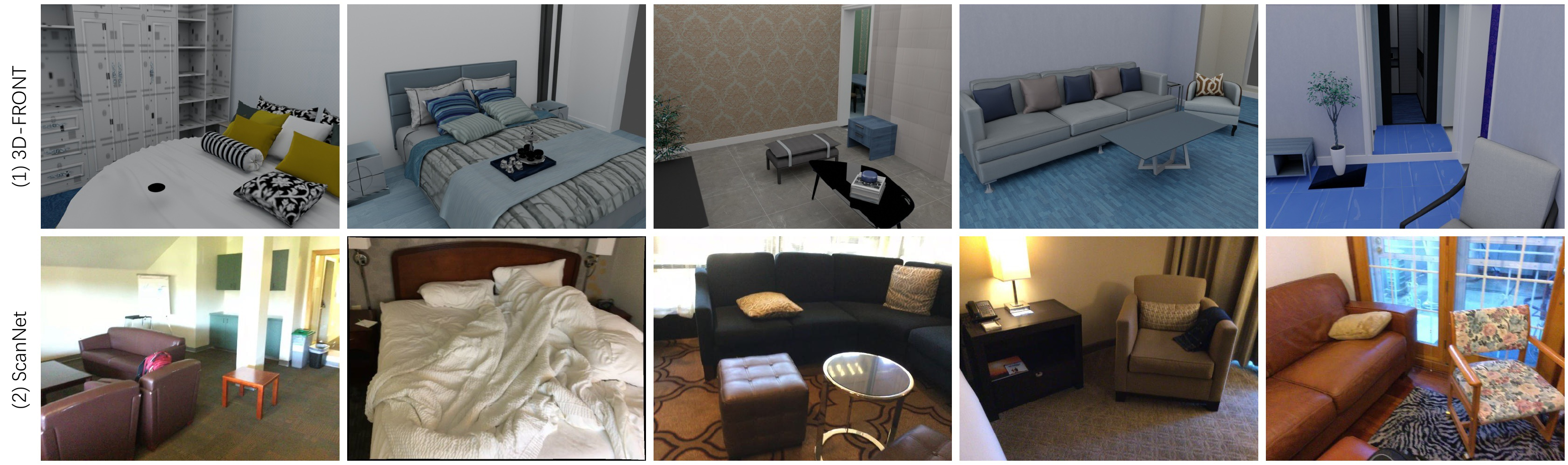}
    \caption{\textbf{Images in 3D-FRONT dataset\cite{fu20213d} and ScanNet dataset\cite{dai2017scannet}}. The first row shows the images in 3D-FRONT dataset. The second row shows the images in ScanNet dataset.}  
    \label{fig: appendix-dataset}
\end{figure*}

\subsection{3D-FRONT}
Here we describe how we preprocess 3D-FRONT \cite{fu20213d} dataset.
First, we randomly pick 88 rooms labeled as living room or bedroom from the dataset.
For each sampled room, we iteratively select 200 camera views and we need to ensure that there is a certain overlap but also distance between the different selected camera views.
The overlap and distance are calculated between the currently sampled camera view and the previously sampled camera views.

Formally, let $\mathbf{K}$, $\mathbf{E} = [\mathbf{R}, \mathbf{t}]$ denote the camera intrinsic and extrinsic parameters respectively.
We use fixed intrinsic for all camera views and only need to sample the extrinsic parameters for each camera view.
The overlap between camera $\mathbf{E}_1$ and camera $\mathbf{E}_2$ is obtained by calculating the Intersection Over Union (IoU) of the two camera frustums
\begin{equation}
    \mathcal{O}(\mathbf{E}_1, \mathbf{E}_2) = \frac{A \cap B}{A \cup B},
\end{equation}
where $A$ and $B$ are the frustums of $\mathbf{E}_1$ and $\mathbf{E}_2$ respectively.
The distance between camera $\mathbf{E}_1$ and camera $\mathbf{E}_2$ is calculated from the camera position and orientation
\begin{equation}
    \mathcal{D}(\mathbf{E}_1, \mathbf{E}_2) = \| \mathbf{t}_1 - \mathbf{t}_2 \|_2 + \arccos( (\text{Tr}( \mathbf{R}_2^{\top} \mathbf{R}_1 ) - 1) / 2 ).
\end{equation}
Suppose we have sampled a series of camera extrinsics $\mathcal{E} = \{ \mathbf{E}_i = [\mathbf{R}_i, \mathbf{t}_i] \}_{i=1}^N$, where $N$ is the number of sampled extrinsics. 
For the camera extrinsic $\mathbf{E}$, We calculate the overlap and distance between $\mathbf{E}$ and $\mathcal{E}$ by the following formulas
\begin{equation}
    \textbf{Overlap}(\mathbf{E}, \mathcal{E}) = \max(\mathcal{O}(\mathbf{E},\mathbf{E}_1), ..., \mathcal{O}(\mathbf{E},\mathbf{E}_N)),
\end{equation}
and
\begin{equation}
    \textbf{Distance}(\mathbf{E}, \mathcal{E}) = \min(\mathcal{D}(\mathbf{E},\mathbf{E}_1), ..., \mathcal{D}(\mathbf{E},\mathbf{E}_N)).
\end{equation}
Once the overlap and distance reach a certain threshold, we add $\mathbf{E}$ to $\mathcal{E}$.
Finally, the camera view selecting algorithm is shown in Alg.\ref{alg: view}.
We show some images of 3D-FRONT\cite{fu20213d} as shown in Fig.\ref{fig: appendix-dataset}.

\begin{algorithm}
	\renewcommand{\algorithmicrequire}{\textbf{Input:}}
	\renewcommand{\algorithmicensure}{\textbf{Output:}}
	\caption{Camera view selecting}
	\label{alg: view}
	\begin{algorithmic}[1]
		\STATE Initialization: $\mathcal{E} \leftarrow \{\ \}, n \leftarrow 0, N_v \leftarrow 200, T_o \leftarrow 0.3, T_d \leftarrow 0.2$
        \REPEAT
        \STATE Sample camera view as $\mathbf{E}$
        \IF{$n = 0$} 
            \STATE $\mathcal{E} \leftarrow \mathcal{E} + \{\mathbf{E}\}$
            \STATE $n \leftarrow n + 1$
        \ELSE
            \STATE $\text{overlap} \leftarrow \textbf{Overlap}(\mathbf{E}, \mathcal{E})$
            \STATE $\text{distance} \leftarrow \textbf{Distance}(\mathbf{E}, \mathcal{E})$
            \IF{ $\text{overlap} \geq T_o$ \AND $\text{distance} \geq T_d$ } 
                \STATE $\mathcal{E} \leftarrow \mathcal{E} + \{\mathbf{E}\}$
                \STATE $n \leftarrow n + 1$
            \ENDIF
        \ENDIF
        \UNTIL $n \geq N_v$
		\ENSURE  A list of camera views $\mathcal{V}$.
	\end{algorithmic}  
\end{algorithm}

\subsection{ScanNet}

In this paper, we randomly select 8 scenes of ScanNet\cite{dai2017scannet} as our test datasets.
The test scene numbers are `scene0204', `scene0205', `scene0269', `scene0289', `scene0456', `scene0549', `scene0587', and `scene0611', respectively.
We also show some images of ScanNet\cite{dai2017scannet} as shown in Fig.\ref{fig: appendix-dataset}.

\subsection{Other Benchmark Datasets}

During training, depth information is needed to determine the selection of positive pairs in our GeoContrast (Sec.\ref{sec:geocontrast}).
Since DTU dataset\cite{jensen2014large} and LLFF dataset\cite{mildenhall2019local, wang2021ibrnet} only contain RGB images and not depth, we use COLMAP\cite{schonberger2016pixelwise} to estimate the depth for each RGB image following \cite{liu2022neural}.
Note that depth information is only used during training, not during testing.

\section{Network Architecture}
\label{app: network}

\begin{figure}
    \centering
    \includegraphics[width=0.9\linewidth]{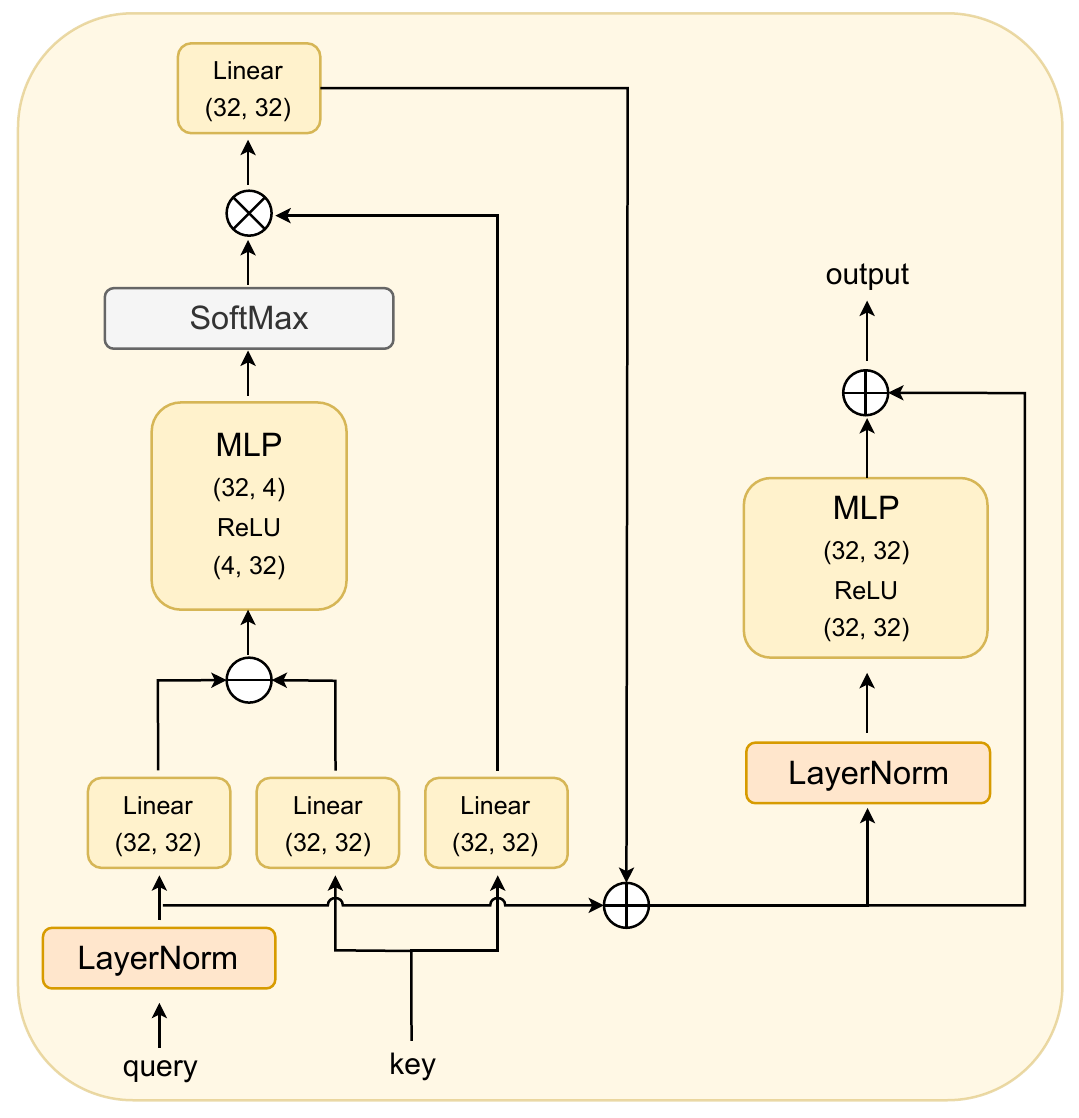}
    \caption{\textbf{Architecture of cross-view attention.}}
    \label{fig:crossview}
\end{figure}

In the feature extraction, we use a U-Net like network, where ResNet34\cite{he2016deep} truncated after layer3 as the encoder, and two additional up-sampling layers with convolutions and skip-connections as the decoder, to extract features from input images following \cite{liu2022neural, wang2021ibrnet}.
All convolution layers use ReLU as activation function and all batch normalization layers are replaced by instance normalization layers.
The output dimensions of each layer of the encoder are 32, 64, 128 respectively, and the output dimensions of each layer of the decoder are 64 and 32 respectively.
As for cross-view attention, we use the subtraction attention\cite{zhao2021point} as our attention module for its effectiveness in geometric relationship reasoning, and the attention layer of the different stages (see Sec.\ref{sec:crossgam}) does not share parameters.
We apply cross-view attention between the encoder and the decoder of U-Net and sample 16 projections as the key values for each query in the first stage, due to GPU memory limitation.
Before entering the cross-view attention, we use a linear layer to reduce the feature dimension from 128 to 32.
After cross-view attention, we also use a linear layer to increase the feature dimension from 32 to 128.
The architecture of cross-view attention is illustrated in Fig.\ref{fig:crossview}.
We implement the rendering network mainly following IBRNet\cite{wang2021ibrnet}, where the multi-view feature aggregation module aggregates the density information of all samples on the ray to enable visibility reasoning, and the ray transformer is then applied to calculate the volume density.

\section{More Experimental Results}
\label{app: experiment}

\subsection{Additional Ablation Study}

\paragraph{Ablation on the number of negative pairs.}
The number of negative pairs is shown to have a larger effect on the performance of contrastive learning as described in \cite{chen2020simple}, and a larger number of negative pairs has a significant advantage over the smaller ones.
Here in our method, we conduct ablation studies with different numbers of negative pairs to see the effect of the number of negative pairs on the model performance.
As shown in Tab.\ref{tab: ablation-neg}, the performance of our method increases as the number of negative pairs increases, with the best performance when the number reaches 512, which is consistent with the phenomenon in \cite{chen2020simple}.
We also tried another way of multi-view consistency optimization, that is, we directly optimize the similarity between positive pairs $\| \mathbf{p}' - \mathbf{q}_+' \|_2$, where $\mathbf{p}'$ and $\mathbf{q}_+'$ are defined in Eq.\ref{eq:weight-cl}.
The result is shown in row 1 of Tab.\ref{tab: ablation-neg}.
We can see that the above optimization method performs worse than our GeoContrast, which further reflects the importance of negative pairs.

\begin{table}
  \centering
  \scalebox{0.85}{\begin{tabular}{cccc}
    \toprule
    Description & PSNR $\uparrow$ & SSIM $\uparrow$ & LPIPS $\downarrow$ \\
    \midrule
    No negative pairs & 23.81 & 0.812 & 0.350 \\
    \midrule
    negative pairs $N_{neg}=32$ & 23.88 & 0.814 & 0.348 \\
    negative pairs $N_{neg}=64$ & 24.19 & 0.819 & 0.342 \\
    negative pairs $N_{neg}=128$ & 24.56 & 0.825 & 0.337 \\
    negative pairs $N_{neg}=256$ & 24.73 & 0.828 & 0.335 \\
    negative pairs $N_{neg}=512$ & \textbf{24.81} & \textbf{0.831} & \textbf{0.333} \\
    negative pairs $N_{neg}=1024$ & 24.80 & \textbf{0.831} & 0.334\\
    negative pairs $N_{neg}=2048$ & 24.79 & 0.830 & \textbf{0.333}\\
    \bottomrule
  \end{tabular}}
  \caption{\textbf{Ablation study on the ScanNet dataset\cite{dai2017scannet} with respect to the number of negative pairs}.}
  \label{tab: ablation-neg}
\end{table}

\paragraph{Ablation on proportion of real data.}

\begin{figure}[H]
    \centering
    \begin{subfigure}{0.48\linewidth}
        \includegraphics[width=\textwidth]{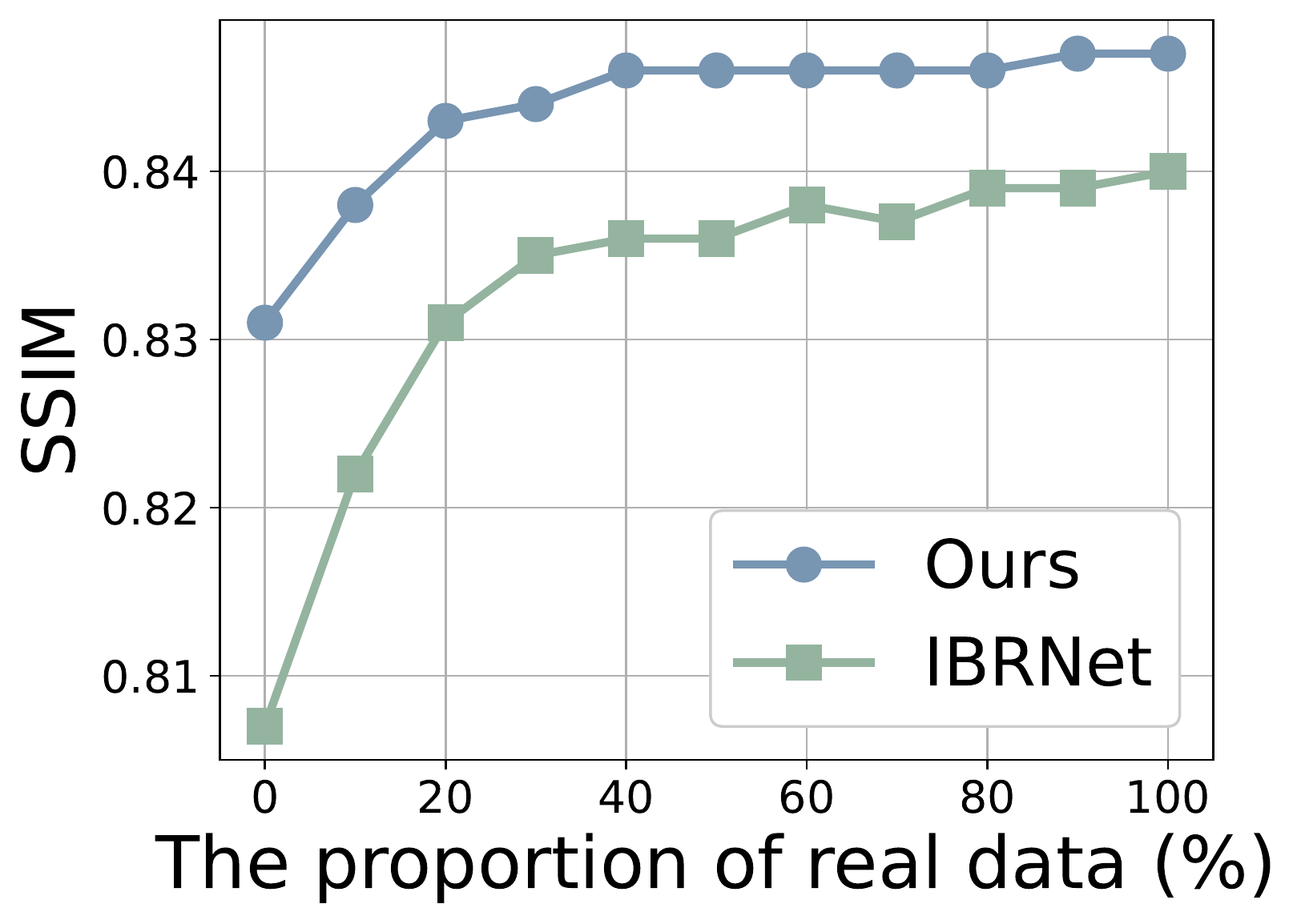}
        \caption{SSIM.}
        \label{fig:psnr}
    \end{subfigure}
    \begin{subfigure}{0.48\linewidth}
        \includegraphics[width=\textwidth]{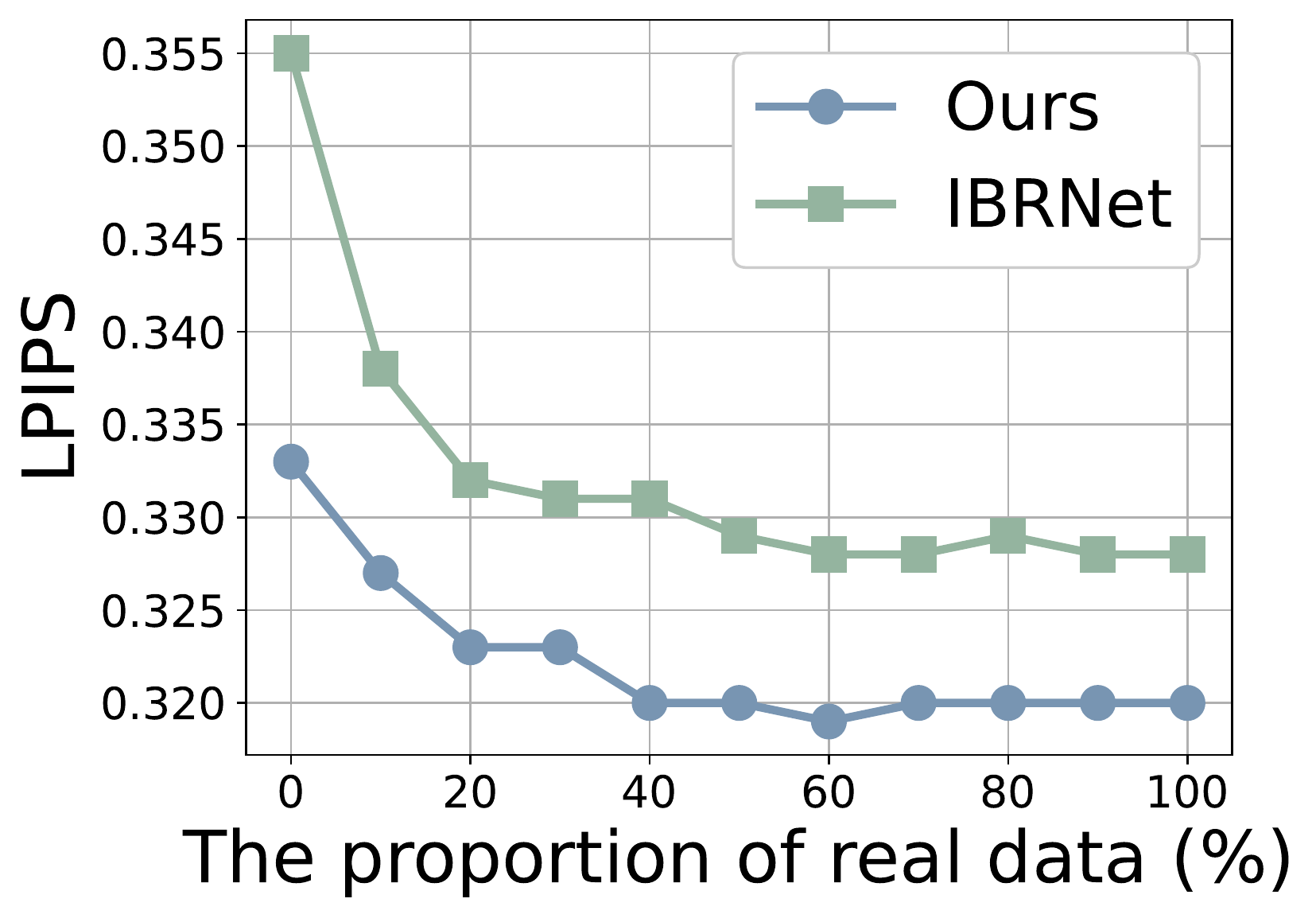}
        \caption{LPIPS.}
        \label{fig:ssim}
    \end{subfigure}
    \caption{\textbf{Curves of SSIM} \textbf{and LPIPS} \textbf{of our method and IBRNet\cite{wang2021ibrnet} with different proportions of real and synthetic data.}}
    \label{fig: propotion2}
\end{figure}

Here we present the SSIM and LPIPS under varying proportions of real and synthetic data as shown in Fig.\ref{fig: propotion2}.
Similar to the curve of PSNR in Fig.\ref{fig:propotion}, the performance of our method continues to improve as the proportion of real data increases, but the performance saturates when the proportion of real data reaches 40\%.

\subsection{NeRF Synthetic}

We also conduct experiments on the NeRF synthetic dataset\cite{mildenhall2021nerf}.
The NeRF synthetic dataset contains 8 objects, each of which has 100 training views and 200 test views at 800 × 800 resolution.
The experimental settings are the same as those on the DTU dataset\cite{jensen2014large} and LLFF dataset\cite{mildenhall2019local}, where we use the Google Scanned Object datqset\cite{downs2022google}, forward-facing datasets\cite{mildenhall2019local, wang2021ibrnet}, and DTU dataset\cite{jensen2014large} as the training dataset.
The quantitative and qualitative results of NeRF synthetic dataset are shown in Tab.\ref{tab: blender} and Fig.\ref{fig: appendix-other} respectively.
Compared with the baseline IBRNet\cite{wang2021ibrnet}, our method has a large performance improvement on NeRF synthetic dataset.
Compared with the state-of-the-art methods\cite{johari2022geonerf, liu2022neural}, our method also achieves comparable results.
The reason why our method underperforms the state-of-the-art methods on the NeRF synthetic dataset may be that the multi-view consistency of synthetic data is relatively better than that of real data, resulting in limited improvement of the methods of learning multi-view consistent representations on synthetic data.
On the other hand, the input to these state-of-the-art methods contains additional geometric information (such as depth) during testing, which will facilitate the modeling of the geometry, while our method's input only contains RGB images.

\begin{table}
  \centering
  \scalebox{0.85}{\begin{tabular}{cccc}
    \toprule
    Method & PSNR $\uparrow$ & SSIM $\uparrow$ & LPIPS $\downarrow$ \\
    \midrule
    PixelNeRF\cite{yu2021pixelnerf} & 22.65 & 0.808 & 0.202 \\
    IBRNet\cite{wang2021ibrnet} & 26.73 & 0.908 & 0.101 \\
    MVSNeRF\cite{chen2021mvsnerf} & 25.15 & 0.853 & 0.159 \\
    GeoNeRF\cite{johari2022geonerf} & \textbf{28.33} & \textbf{0.938} & \textbf{0.060} \\
    Neuray\cite{liu2022neural} & 28.29 & 0.927 & 0.080 \\
    Ours & 27.92 & 0.930 & 0.078\\
    \bottomrule
  \end{tabular}}
  \caption{\textbf{Quantitative comparisons on the NeRF Synthetic dataset\cite{mildenhall2021nerf}}.}
  \label{tab: blender}
\end{table}

\subsection{Additional Qualitative Results}

\begin{figure*}
    \centering
    \includegraphics[width=0.95\textwidth]{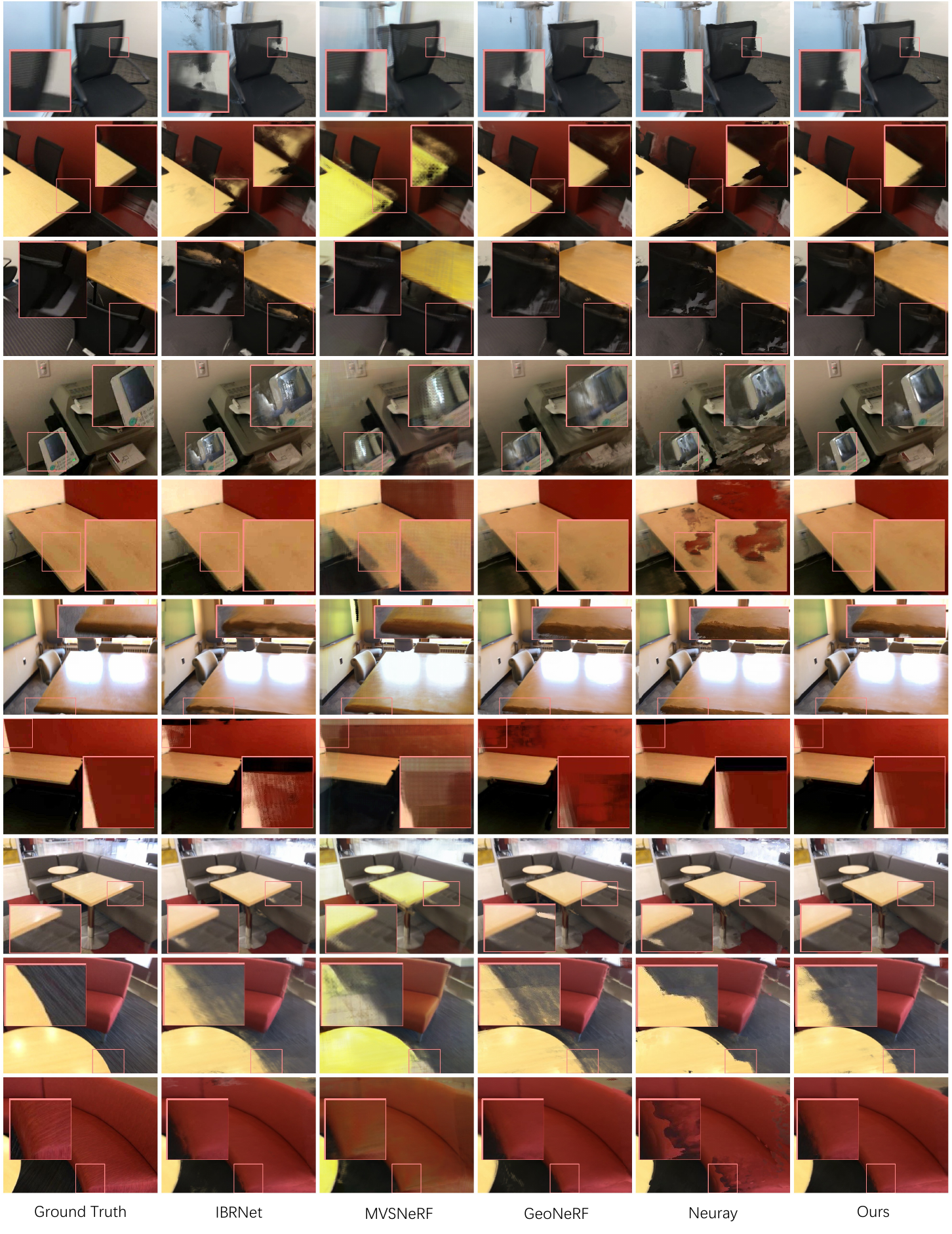}
    \caption{\textbf{Qualitative comparison on ScanNet dataset\cite{dai2017scannet}}. The first column shows the ground truth images. The last column shows the rendered images of our method. The remaining columns represent the images rendered by IBRNet\cite{wang2021ibrnet}, MVSNeRF\cite{chen2021mvsnerf}, GeoNeRF\cite{johari2022geonerf}, Neuray\cite{li2021neural}, respectively. Each model is trained on the synthetic dataset.}  
    \label{fig: appendix-scannet}
\end{figure*}

\begin{figure*}
    \centering
    \includegraphics[width=0.93\textwidth]{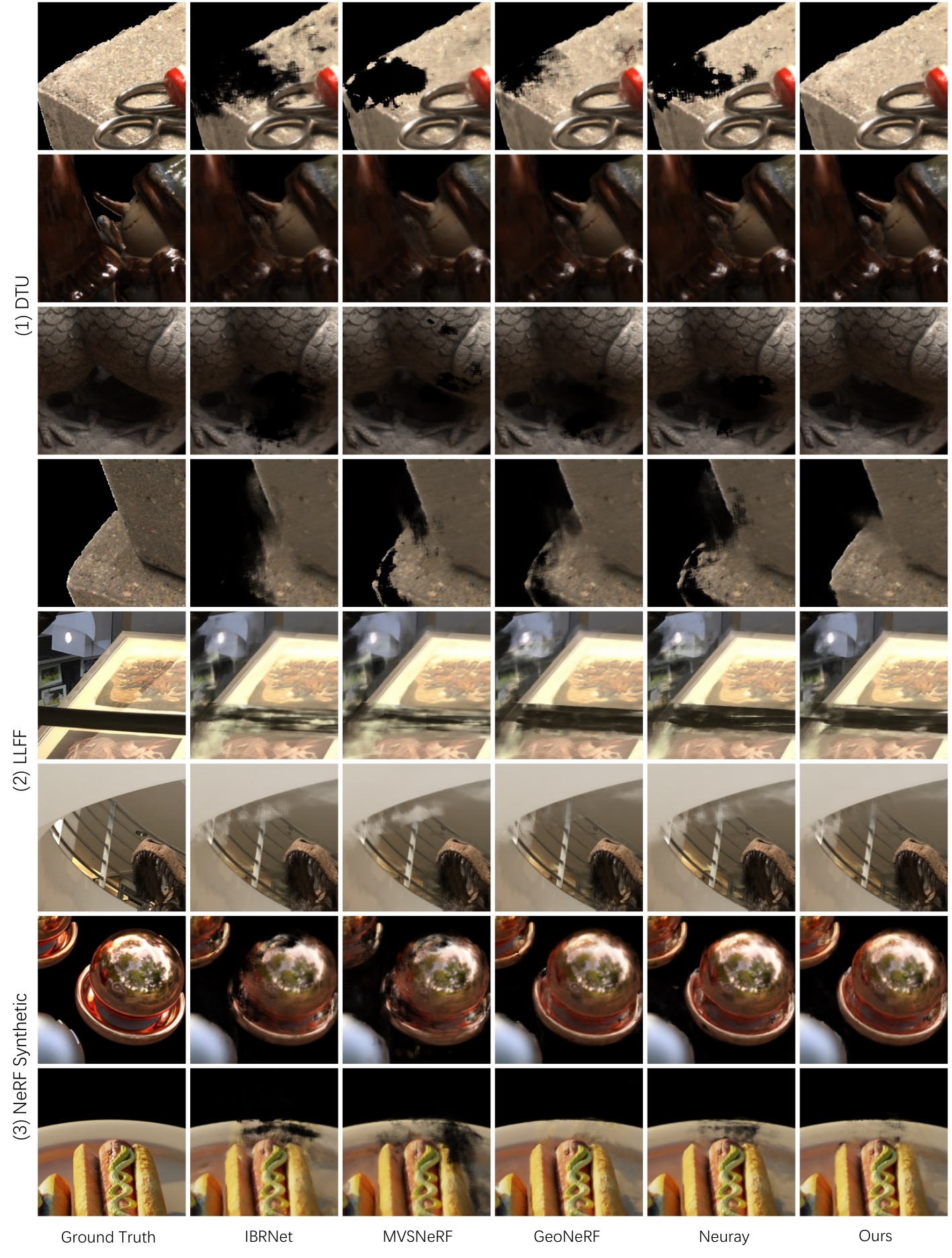}
    \caption{\textbf{Qualitative comparison on DTU dataset\cite{jensen2014large} (rows 1 to 4), LLFF dataset\cite{mildenhall2019local} (rows 5 to 6), and NeRF synthetic dataset\cite{mildenhall2021nerf} (rows 7 to 8)}. The first column shows the ground truth images. The last column shows the rendered images of our method. The remaining columns represent the images rendered by IBRNet\cite{wang2021ibrnet}, MVSNeRF\cite{chen2021mvsnerf}, GeoNeRF\cite{johari2022geonerf}, Neuray\cite{li2021neural}, respectively. }  
    \label{fig: appendix-other}
\end{figure*}

In this section, we provide additional qualitative results.
Fig.\ref{fig: appendix-scannet} shows the qualitative results of IBRNet\cite{wang2021ibrnet}, MVSNeRF\cite{chen2021mvsnerf}, GeoNeRF\cite{johari2022geonerf}, Neuray\cite{liu2022neural} and our method on ScanNet dataset\cite{dai2017scannet}.
All models are trained on 3D-FRONT\cite{fu20213d} and the experimental settings are the same as in Sec.\ref{sec:dataset}.
Fig.\ref{fig: appendix-other} shows the qualitative results on DUT dataset\cite{jensen2014large}, LLFF dataset\cite{mildenhall2019local}, and NeRF synthetic dataset\cite{mildenhall2021nerf}, and the experimental settings are the same as in Sec.\ref{sec: other-data}.

\section{Limitation and Failure Case}
\label{app: failure}

\begin{figure*}
    \centering
    \includegraphics[width=0.99\textwidth]{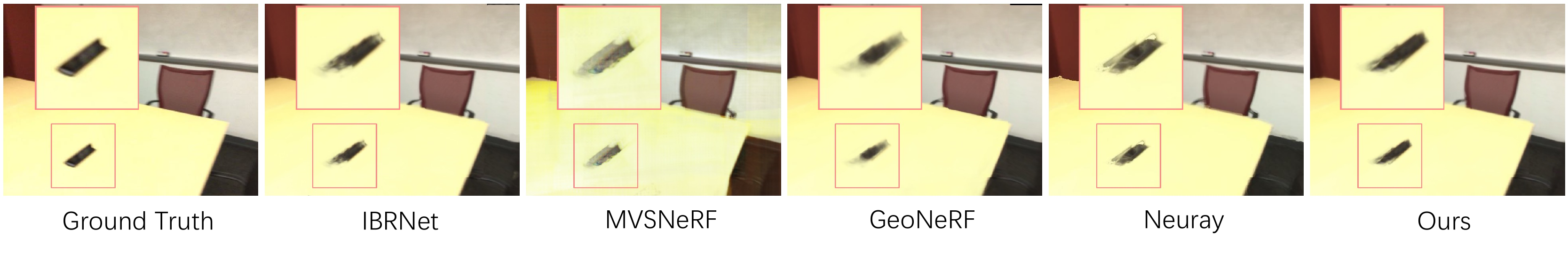}
    \caption{\textbf{Failure case on ScanNet dataset\cite{dai2017scannet}}. The first column shows the ground truth images. The last column shows the rendered images of our method. The remaining columns represent the images rendered by IBRNet\cite{wang2021ibrnet}, MVSNeRF\cite{chen2021mvsnerf}, GeoNeRF\cite{johari2022geonerf}, Neuray\cite{li2021neural}, respectively. Each model is trained on the synthetic dataset.}  
    \label{fig: appendix-failure}
\end{figure*}

Our method generally achieves high-quality image rendering under the synthetic-to-real setting.
However, previous generalizable NeRF methods\cite{wang2021ibrnet, chen2021mvsnerf, johari2022geonerf, liu2022neural}, as well as ours, struggle to generate high-quality images for highly blurred scenes, which are frequently found in real dataset.
We show an example in Fig.\ref{fig: appendix-failure}, where motion blur occurs in the pink boxed region and all methods fail to predict a sharp image.
Deblur-NeRF\cite{ma2022deblur} tries to recover a sharp NeRF from blurry input with the Deformable Sparse Kernel module.
However, Deblur-NeRF only considers the per-scene optimization case.
Rendering images with high bulr undering the synthetic-to-real generalization setting is a challenging problem and it can be an interesting and practical future direction.

\end{document}